\documentclass[journal]{IEEEtran}
\usepackage{ amssymb   }
\usepackage{ mathdots  }
\usepackage{ amsmath   }
\usepackage{ color     }
\usepackage{ graphicx  }
\usepackage{ amsfonts  }
\usepackage{ amssymb   }
\usepackage{ epstopdf  }
\usepackage{ upgreek   }
\usepackage{ bm        }
\usepackage{ graphicx  }
\usepackage{ subfigure }
\usepackage{ mathtools }
\usepackage{ algorithm }
\usepackage{algorithmic}
\usepackage{ booktabs  }
\usepackage{ multirow  }
\usepackage{ caption   }
\usepackage{ array     }
\usepackage{ float     }
\usepackage{   soul    }
\usepackage[normalem]{ulem}

\usepackage[final]{pdfpages}
\usepackage[noadjust]{cite}
\newcolumntype{L}[1]{>{\raggedright\let\newline\\\arraybackslash\hspace{0pt}}m{#1}}
\newcolumntype{C}[1]{>{\centering\let\newline\\\arraybackslash\hspace{0pt}}m{#1}}
\newcolumntype{R}[1]{>{\raggedleft\let\newline\\\arraybackslash\hspace{0pt}}m{#1}}

\DeclareMathOperator*{\argmin}{arg\,min}

\newcommand\norm[1]{\left\lVert#1\right\rVert}
\newcommand{\bs}{\boldsymbol{s}}
\newcommand{\ba}{\boldsymbol{a}}
\newcommand{\bw}{\boldsymbol{w}}
\newcommand{\btheta}{\boldsymbol{\theta}}

\definecolor{antiquefuchsia}{rgb}{0.57, 0.36, 0.51}
\definecolor{cadmiumred}{rgb}{0.89, 0.0, 0.13}

\begin{document}
	% ===========================
	%          title
	% ===========================
	\title{Load Balancing for Ultra-Dense Networks: A Deep Reinforcement Learning Based Approach}
	% ===========================
	%          author
	% ===========================
	\author{Yue~Xu,~\IEEEmembership{Student Member,~IEEE,}
		Wenjun~Xu,~\IEEEmembership{Senior Member,~IEEE,}
		Zhi~Wang, Jiaru~Lin,~\IEEEmembership{Member,~IEEE,}
		and~Shuguang~Cui,~\IEEEmembership{Fellow,~IEEE,}
		
		\thanks{The work was supported in part by grants NSFC-61629101, NSFC-61771066, No. ZDSYS201707251409055, No. 2017ZT07X152, No. 2018B030338001, No. 2018YFB1800800, CNS-1824553, DMS-1622433, AST-1547436, and ECCS-1659025. (Co-correspondence Authors: Wenjun Xu and Shuguang Cui.)}
		\thanks{Yue~Xu, Wenjun~Xu, and Jiaru~Lin are with the Key Lab of Universal Wireless Communications, Ministry of Education, Beijing University of Posts and Telecommunications, Beijing, 100876 China (Email:xuy@bupt.edu.cn; wjxu@bupt.edu.cn; jrlin@bupt.edu.cn).}
		\thanks{Zhi~Wang is with the State Key Laboratory of Networking and Switching Technology, Beijing University of Posts and Telecommunications, Beijing, 100876 China. (Email:wangzhi@bupt.edu.cn)}
		\thanks{Shuguang~Cui is with the Shenzhen Research Institute of Big Data and School of Science and Engineering, the Chinese University of Hong Kong, Shenzhen, China, 518172. He has also been affiliated with Department of Electrical and Computer Engineering, University of California, Davis, CA, 95616. (Email: robert.cui@gmail.com).}
	}
	\maketitle
	
	% ============================
	%          abstract
	% ============================
	\begin{abstract}
		In this paper, we propose a deep reinforcement learning (DRL) based mobility load balancing (MLB) algorithm along with a two-layer architecture to solve the large-scale load balancing problem for ultra-dense networks (UDNs).
		Our contribution is three-fold. 
		First, this work proposes a two-layer architecture to solve the large-scale load balancing problem in a self-organized manner. The proposed architecture can alleviate the global traffic variations by dynamically grouping small cells into self-organized clusters according to their historical loads, and further adapt to local traffic variations through intra-cluster load balancing afterwards. 
		Second, for the intra-cluster load balancing, this paper proposes an off-policy DRL-based MLB algorithm to autonomously learn the optimal MLB policy under an asynchronous parallel learning framework, without any prior knowledge assumed over the underlying UDN environments. Moreover, the algorithm enables joint exploration with multiple behavior policies, such that the traditional MLB methods can be used to guide the learning process thereby improving the learning efficiency and stability.
		Third, this work proposes an offline-evaluation based safeguard mechanism to ensure that the online system can always operate with the optimal and well-trained MLB policy, which not only stabilizes the online performance but also enables the exploration beyond current policies to make full use of machine learning in a safe way. 
		Empirical results verify that the proposed framework outperforms the existing MLB methods in general UDN environments featured with irregular network topologies, coupled interferences, and random user movements, in terms of the load balancing performance. 
	\end{abstract}
	% ============================
	%        IEEEkeywords
	
	% ============================
	%        Introduction
	% ============================
	\section{Introduction}
	% 5G UDN and Load balancing
	The fifth generation (5G) system is anticipated to meet the explosive growth of mobile data traffic, where the ultra-dense network (UDN) has been widely recognized as one of the most promising solutions~\cite{8387197, 8387203}. In UDNs, the small cell base stations (SBSs) are densely deployed to reduce the distance between the mobile users and the access points, such that the link quality could be substantially improved to increase the network capacity.
	However, such densification exacerbates the issue of traffic fluctuation compared with the system with large cells, which will greatly affect the network performances, e.g., the quality of service (QoS) experienced by users~\cite{8436053, 8334688}.
	Therefore, load balancing, which aims at alleviating traffic fluctuation, is becoming increasingly important in UDNs.
	On the other hand, the ultra-dense deployment of small cells brings numerous challenges in large-scale network control, communications and optimizations, where configuring the network for load balancing in UDNs is different from that in traditional cellular networks~\cite{8436053, 8334688}. 
	For example, the traditional load balancing in cellular networks through power control~\cite{1379093} or changing radiation patterns~\cite{4524857} may change the association of all the users in the coverage area, which could downgrade the performance due to the severely coupled interference in a UDN with dense and irregular small cell topology. 
	Thus, exploring the appropriate manner for load balancing in 5G UDNs to accommodate its irregular network topologies, complicated interference relationships, and diversified user mobility patterns still remains a challenging research problem.
	
	Recently, self-organizing networks (SONs), which promote autonomic learning and self-management, have already been recognized as an attractive paradigm to realize adaptive management for 5G UDNs in a scalable manner~\cite{8506421, 8337738}. 
	Therefore, it is promising to incorporate self-organizing functionalities into 5G UDNs to enable adaptive and scalable load balancing.
	The traditional load balancing technique with self-organizing characteristics is referred to as mobility load balancing (MLB)~\cite{6678648, 5594565, 6398873, 8094955}, which usually optimizes the load distribution via tuning a parameter named cell individual offset (CIO) to control the user handovers. 
	Current literature has proposed various MLB methods. 
	% Rule-based controllers
	For example, the rule-based methods map system states into pre-defined load balancing rules. Raymond \textit{et al.}~\cite{5594565} proposed to change the CIOs via the rule with a fixed step-size based on the load difference between adjacent cells; Yang \textit{et al.}~\cite{6398873} later proposed to use the rule with an adaptive step-size to reach load equilibrium more efficiently.
	However, for rule-based controllers, the predefined rules are expected to cover all possible circumstances, which requires a perfect understanding on both the system and the rules.  
	% Game-theoretic models
	Game-theoretic methods, on the other hand, model the MLB process as an ongoing game among cells. For example, Sheng \textit{et al.}~\cite{6678648} modeled the MLB problem as a Cournot game; Park \textit{et al.}~\cite{8094955} optimized the user assignment and the target cell selection based on the Kuramoto synchronization and matching theory, respectively. 
	However, the equilibrium and optimality of game-theoretic models are usually built upon specific assumptions over the wireless gaming environment. 
	
	% Reinforcement learning based models
	On the other hand, the emerging paradigm of machine learning and big data is reshaping the future of wireless networks~\cite{7295483, chen2017machine, 8304392, 8387203}. 
	Reinforcement learning (RL), which is one class of machine learning methods that can adapt to unknown environments by learning from the environmental feedbacks, has already been applied in load balancing~\cite{6384872, 7393587}.
	However, the existing RL-based load balancing methods need to first quantize the system states and the control actions to construct a tabular environment~\cite{6384872, 7393587}, which suffers from the curse of dimensionality, i.e., the numbers of actions and states increase exponentially with the degrees of freedom, thus largely hindering their applications to the complex 5G UDN environments. 
	Noticeably, the recent combination of deep learning and RL, coined as DRL, has greatly improved the generalization capability of RL to achieve the state-of-the-art performances in many fields, such as competitive gamings~\cite{silver2016mastering, silver2017mastering}, biological data analysis~\cite{8277160}, mobility robustness optimization~\cite{8387430}, etc. Moreover, our previous study indicates that the DRL can also achieve state-of-art load balancing performance under a small-scale SON~\cite{Xu1905:Deep}.
	These successes envision a bright future to exploit DRL in realizing adaptive and autonomous large-scale load balancing under complex UDN environments, e.g., coupled interferences and irregular network topologies. However, this paradigm still remains to be explored.
	
	\subsection{Contributions}
	In this paper, we investigate the large-scale MLB problem in UDNs with emphasis on three important features, i.e., generalization, adaptation, and scalability.
	The main contributions are summarized as follows.
	\begin{itemize}
		\item \textbf{Architecture:} This work proposes a two-layer MLB architecture to handle the large-scale load balancing problem for UDNs in a self-organized manner. The top layer aims at dynamically grouping the underlying SBSs into self-organized clusters with adaptation to global traffic variations by using a tailored k-means algorithm, which picks the top overloaded SBSs as the initial cluster centroids and then groups SBSs based on their locations; the bottom layer aims at balancing the intra-cluster load distribution with adaptation to local traffic variations by using a DRL-based algorithm. Such a self-organized mechanism breaks the large-scale load balancing problem into smaller pieces that are easier to be handled distributively, thereby improving the system scalability and cost-efficiency.
		
		\item \textbf{Algorithm:} First, this work proposes an off-policy DRL-based algorithm for MLB, which approximates the MLB policy with deep neural networks to largely improve the model generalization capability over complex UDN environments. The algorithm can be trained under an asynchronous parallel learning framework to autonomously learn the optimal MLB policy without any prior knowledge over the underlying wireless environments. 
		Second, the proposed algorithm enables joint exploration with multiple behavior policies, such that existing MLB methods could be used to guide the learning process thereby improving the learning efficiency and stability at the early stage.
		Empirical results verify that the proposed DRL-based MLB model can adapt to general UDN environments featured with irregular network deployment, coupled interferences, and random user mobility patterns, and outperforms existing MLB methods considerably.
		
		\item \textbf{Mechanism:} This work proposes an offline-evaluation based safeguard mechanism to improve the online control performance in practical systems. 
		Specifically, the proposed mechanism runs an online branch for system control and an offline branch for policy learning in parallel. The online branch always uses the best MLB policy well-trained by the offline branch. 
		As such, the mechanism can avoid the risk of exhibiting unstable or even destructive behaviors when following the under-trained policy at the early learning stage.
		Moreover, such a mechanism also enables policy exploration in a safe way, which is of vital importance to make full use of machine learning to go beyond existing methods for superior alternatives.
	\end{itemize}
	Indeed, the proposed architecture, algorithm and mechanism form a general autonomous and intelligent network control framework, which is also promising to solve other large-scale network control problems in the future systems by changing the learning context.
	
	The remainder of this paper is organized as follows. 
	Section~\ref{sec:system_model} introduces the two-layer MLB architecture and the system model.
	Section~\ref{sec:top-layer} presents the centralized clustering algorithm.
	Section~\ref{sec:bottom-layer} presents the self-organized DRL-based MLB algorithm.
	Section~\ref{sec:safeguard} presents the safeguard mechanism for online control.
	Section~\ref{sec:results} discusses the simulation results. 
	Finally, Section~\ref{sec:conclusion} concludes the paper.
	
	% ===========================
	%          section
	% ===========================
	%\clearpage
	\section{System Model}
	\label{sec:system_model}
	\subsection{Two-layer MLB Architecture for Load Balancing in UDNs}
	\label{sec:Two-layer}
\begin{figure}[tb]
		\centering
		\includegraphics[trim = 10 10 10 10, clip, width=8cm]{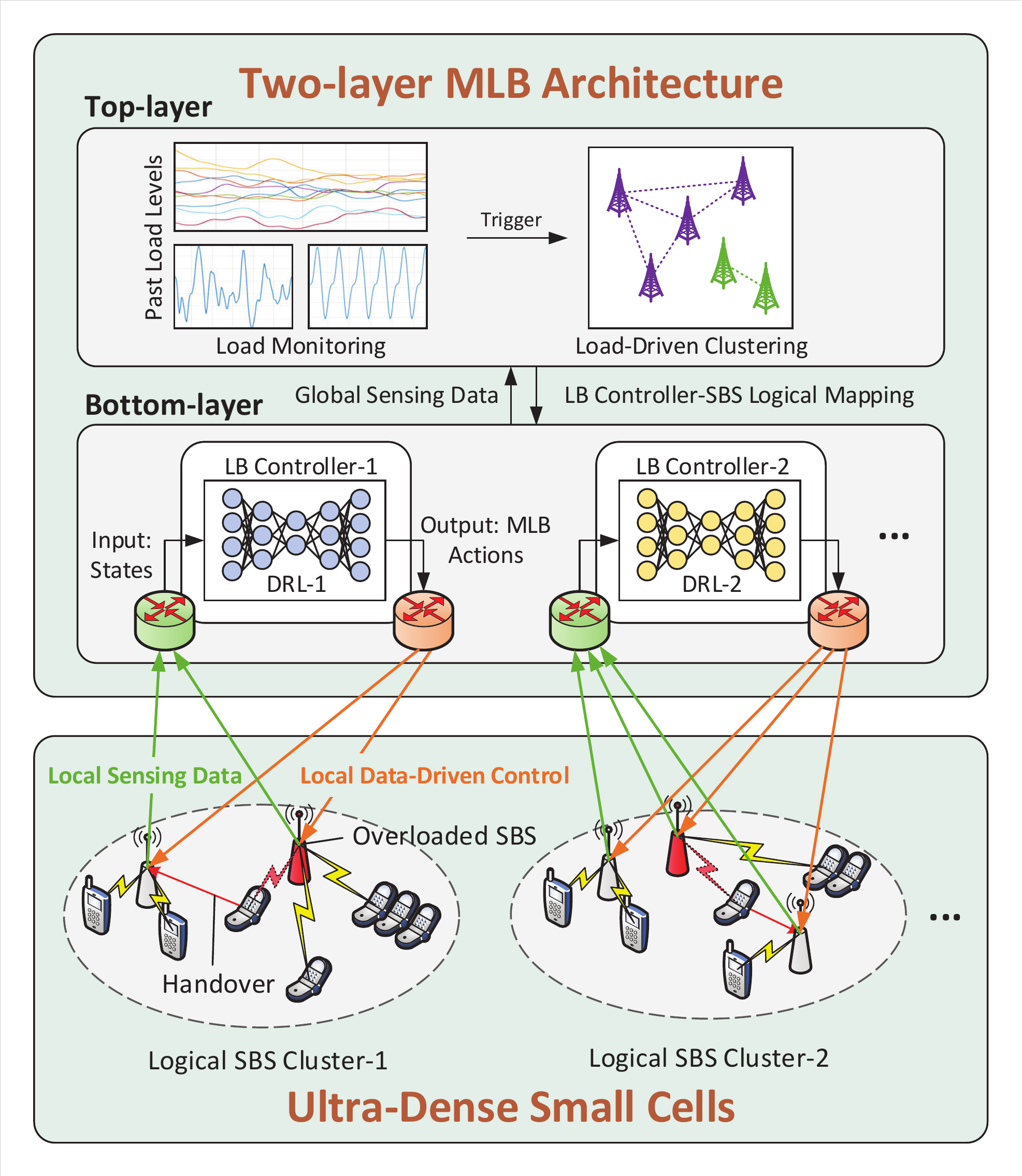}
		\caption{A two-layer MLB architecture for UDNs}
		\label{fig:scenario}
	\end{figure}

	In this paper, we propose a two-layer MLB architecture to control large-scale load balancing for UDNs in a self-organized manner.
	As shown in Fig.~\ref{fig:scenario}, the two-layer architecture includes a centralized load-driven \textit{clustering controller} at the top layer, and multiple self-organized \textit{load balancing controllers} at the bottom layer. 
	The top layer aims at grouping all the SBSs into a bunch of clusters according to their historical load levels, repeatedly performed on the order of hours; the bottom layer aims at balancing the load within each cluster, repeatedly performed on the order of seconds up to minutes.
	In this way, the top layer adapts to the dynamic global traffic fluctuations from a macroscopic view, while the bottom layer tunes the intra-cluster load distribution at a finer level.
	
	The main advantages of the proposed architecture can be summarized as follows.
	\begin{enumerate}
		\item \textbf{Scalability}: The large-scale UDN can now be managed by multiple self-organized controllers, where each of them only needs to work with a small portion of the entire network, i.e., one cluster. As such, each controller only needs to adapt their configurations autonomously based on \textit{partial information} of the network, which largely reduces the global information exchanges and is more robust to the network topology variation, thus having a strong scalability.
		\item \textbf{Efficiency}: The scale of the non-convex load balancing problem increases with the the number of SBSs, which implies an increasing amount of training overhead when using learning-based algorithms. 
		Meanwhile, the global minimum is not easy to obtain.
		The proposed self-organized paradigm breaks the large problem into smaller pieces that are easier to handle, such that the high computational complexity can be distributed to a bunch of self-organized controllers, which can work in parallel. In this way, the algorithm is expected to find better locally optimal solutions with improved cost-efficiency.
	\end{enumerate}
	
\begin{figure*}[thb]
		\centering
		\includegraphics[trim = 10 10 10 10, clip, width=15cm]{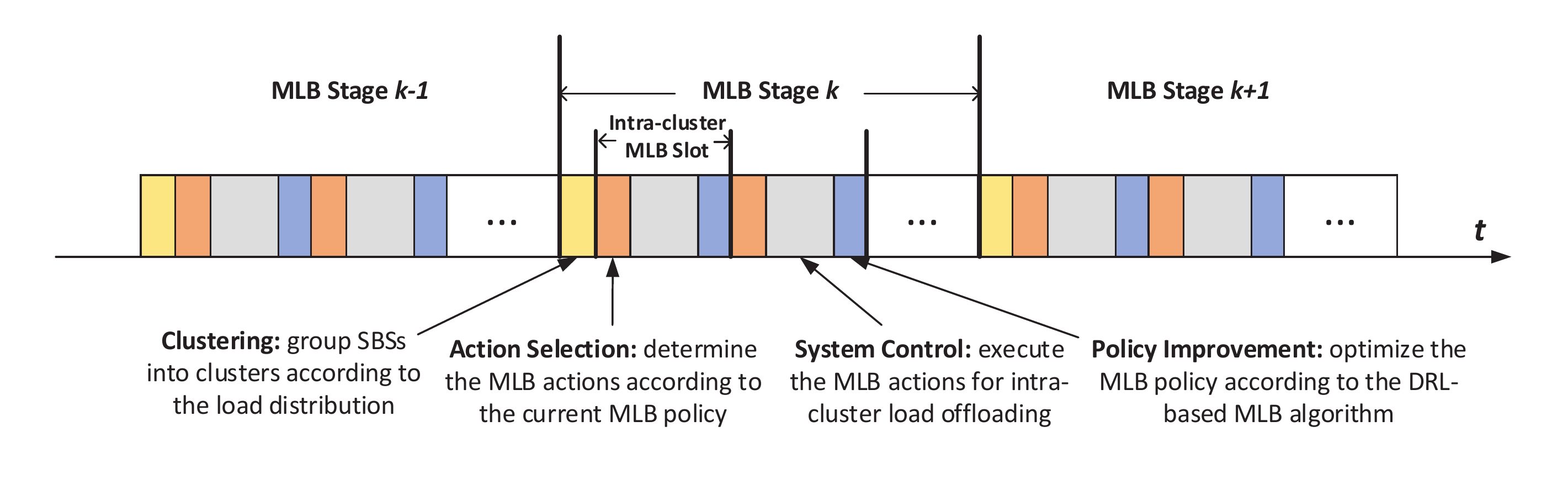}
		\caption{Illustration of the MLB procedure based on the two-layer architecture}
		\label{fig:time_slot}
	\end{figure*}

	In this paper, we consider a MLB scenario where the SBSs are densely and \textit{randomly} deployed. The overloaded SBSs handover the mobile users to their corresponding neighboring SBSs for traffic offloading by tuning the CIOs. The distributed SBSs collect the environment data, e.g., SBS loads, and deliver the data to the controllers at the bottom layer. The top layer dynamically groups the SBS into a bunch of clusters by using a load-driven clustering algorithm; the bottom layer performs self-organized intra-cluster load balancing based on partial network information by using a DRL-based learning algorithm. 
	%Detailed formulations and algorithms are presented in the following sections.
	The detailed work-flow is shown in Fig.~\ref{fig:time_slot} and can be interpreted as follows.
	The clustering at the top layer is triggered either periodically or based on certain traffic fluctuations, e.g., the sudden increase of traffic due to some sport events, the daily tidal effect, etc. 
	We refer to the time period between two clustering operations at the top layer as one \textit{MLB stage}. 
	First, at the beginning of MLB stage $ k $, the top layer groups all the SBSs into multiple clusters according to their averaged load levels during the previous stage. The clustering result stays unchanged over the stage. Then, the bottom layer performs intra-cluster MLB for each cluster by executing: 1) action selection, i.e., selecting the MLB actions according to the DRL-based MLB policy; 2) system control, i.e., executing the control actions for intra-cluster load offloading; 3) policy improvement, i.e., improving its DRL-based MLB policy by learning. The stage $ k $ repeats the above three procedures until the next stage~$ k+1 $, i.e., until the re-clustering is triggered. 
	
	\subsection{Handover Management}
	\subsubsection{Handover}
	The handover process in MLB aims at transferring a mobile user from its serving SBS to a neighboring SBS if better signal quality can be reached.
	Specifically, at a given time $ t $, the handover of one mobile user from its serving SBS $ i $ to a neighboring SBS $ j $ is triggered according to the A3 condition~\cite{3gpppresence} as
	\begin{equation}\label{eq:A3 condition}
	F^t_j - F^t_i > O^t_{ij} + Hys,
	\end{equation}
	where $ F^t_i $ and $ F^t_j $ are the user's reference signal received power (RSRP) of its serving SBS $ i $ and the neighboring SBS $ j $, respectively, $ Hys $ is the handover hysteresis, which is usually a fixed value to prevent frequent handovers, and $ O^t_{ij} $ is the CIO between SBS $ i $ and SBS $ j $. 
	Note that $ O^t_{ij} $ is usually symmetrical, i.e., $ O^t_{ij} = -O^t_{ji} $, to ensure that a user handed over from one SBS to another will not be handed straight back to prevent the ping-pong effect~\cite{5594565}.
	
	According to (\ref{eq:A3 condition}), decreasing a proper value of $ O^t_{ij} $ can trigger user handovers from SBS $ i $ to SBS $ j $, thereby offloading the load from SBS $ i $ to SBS $ j $, and vise versa. 
	Therefore, the key of MLB is to find the best CIO tuning policy to trigger user handovers optimally.
	
	\subsubsection{User Throughput}
	The signal-to-noise-plus-interference ratio (SINR) of user $ u $ in SBS $ i $ at time $ t $ is given by
	\begin{equation}\label{key}
	\text{SINR}^t_{u} = \frac{P_i G^t_{ui}}{N_0 + \sum_{j \in \mathcal{I}, j \neq i}P_j G^t_{uj}},
	\end{equation}
	where $ \mathcal{I} $ is the SBS set, $ P_i $ is the transmission power of SBS $ i $, $ G^t_{ui} $ is the channel power gain between user $ u $ and SBS $ i $ at time $ t $, and $ N_0 $ is the noise power, which is assumed, without loss of generality, to be the same for all the users.
	We assume that the smallest resource unit to allocate is the physical resource block (PRB). 
	For a given user $ u $, the maximum transmission rate of \textit{one} PRB at time $ t $ is given by
	\begin{equation}\label{eq:user_rate}
	R^t_u = B\log_{2}(1 + \text{SINR}^t_{u}),
	\end{equation}
	where $ B $ is the spectrum bandwidth of one PRB.
	
	\subsubsection{Load}
	In this paper, we assume that each user has a constant bit rate (CBR) requirement $ M^t_u $ at time $ t $. The required number of PRBs to meet the demand $ M^t_u $ is
	\begin{equation}\label{key}
	N^t_u = \min\left\lbrace \frac{M^t_{u}}{R^t_{u}}, N_c \right\rbrace ,
	\end{equation}
	where $ N_c $ is a constant threshold to keep the number of PRBs occupied by users with poor channel quality to be under a reasonable level. 
	Finally, the load of SBS $ i $, defined as the ratio of users' required number of PRBs versus the total number of PRBs, can be written as
	\vspace*{-0.5\baselineskip}
	\begin{equation}\label{key}
	\rho^t_i = \frac{\sum_{u \in \mathcal{U}^t_i} N^t_u}{N^p_i},
	\end{equation}
	where $ N^P_i $ is the total number of PRBs of SBS $ i $, and $ \mathcal{U}^t_i $ is the set of users assigned to SBS $ i $ at time $ t $. 
	
	\subsection{Preliminaries on Reinforcement Learning}
	\label{sec:off-policy-RL}
	\subsubsection{Reinforcement Learning}
	In this paper, we formulate the load balancing problem with RL. 
	Specifically, RL aims at maximizing a cumulative reward by selecting a sequence of optimal actions under  different system states in a stochastic unknown environment~\cite{sutton2018reinforcement}. The dynamics is usually modeled as a Markov decision process (MDP), which can be characterized by a state space $ \mathcal{S} $, an action space $ \mathcal{A} $, a reward function $ r : \mathcal{S} \times \mathcal{A} \rightarrow \mathbb{R}^1 $, and a stationary transition probability satisfying the Markov property $ p(\bs_{t+1}|\bs_1, \ba_1, \cdots, \bs_t, \ba_t) = p(\bs_{t+1}|\bs_t, \ba_t) $, where $ \bs \in \mathcal{S}, \ba \in \mathcal{A} $.
	Specifically, at each state $ \bs_t \in \mathcal{S} $, the RL agent selects an action $ \ba_t \in \mathcal{A} $ by following a policy $ \pi$ to interact with the environment, and receives a reward $ r(\bs_t, \ba_t) $; then the state $ s_t $ moves on to $ \bs_{t+1} $ and the system dynamics starts the next round. 
	
	\subsubsection{Value Function}
	In RL, the value function for a given policy $ \pi $  at state $ \bs $ is defined to be the received long-term expected cumulative rewards starting at state $ \bs $ and following the policy $ \pi $ thereafter~\cite{sutton2018reinforcement}, i.e.,
	\begin{equation}\label{key}
	V^{\pi}(\bs) = \mathbb{E}^{\pi}\left[ \sum_{t=0}^{\infty} \gamma^t r(\bs_t, \ba_t) \Big| \bs_t=\bs \right],
	\end{equation}
	where $ \gamma \in [0, 1] $ is the discount factor. 
	Similarly, the Q-function for a given policy $ \pi $ when choosing action $ \ba $ at state $ \bs $ is defined as  
	\begin{equation}\label{key}
	Q^{\pi}(\bs, \ba) = \mathbb{E}^{\pi}\left[ \sum_{t=0}^{\infty} \gamma^t r(\bs_t, \ba_t) \Big| \bs_t=\bs, \ba_t=\ba \right].
	\end{equation}
	Generally, the goal of RL is to find the optimal policy $ \pi^* $ which maximizes the cumulative discounted rewards from the start state $ \bs_0 $~\cite{sutton2018reinforcement}. 
	
	\subsubsection{Off-policy Learning}
	The learning and exploration in RL usually involve two policies: the \textit{target policy} and the \textit{behavior policy}. The target policy is optimized by using the samples generated by following the behavior policy. In practice, the target policy is usually used for system control, while the behavior policy is used for exploration.
	In the literature, there are two classes of learning methods in RL~\cite{sutton2018reinforcement}, i.e., the \textit{on-policy} methods and the \textit{off-policy} methods. 
	The on-policy methods use the same policy as the target policy and the behavior policy, in other words, using the same policy for both system control and exploration.
	The off-policy methods, on the other hand, separate the target policy from the behavior policy, where the target policy is optimized by using the samples generated from the behavior policy based on the importance sampling technique~\cite{sutton2018reinforcement}.
	
	In this paper, we focus on the off-policy RL for MLB with the following motivations.
	First, it would be problematic or even dangerous to apply the on-policy methods to online systems, since it requires online exploration, i.e., performing random actions for system control in order to generate randomized learning samples. Comparatively, with off-policy learning, we can employ a deterministic policy for online control while following a different stochastic behavior policy to explore learning samples offline, e.g., by using a network evaluation system, thus avoiding the risk of online exploration.
	Second, the samples generated from different behavior policies or in previous explorations can be reused in the off-policy RL, which is the foundation for our later proposed learning scheme under multiple behavior policies.

	\subsection{Problem Formulation}
	\label{sec:problem_formulation}
	The clustering at the top layer partitions the large-scale load balancing problem into multiple small intra-cluster load balancing problems, where each load balancing controller at the bottom layer only needs to tune the CIOs of its corresponding intra-cluster SBSs.
	
	In this paper, we model the MLB problem in UDN as a MDP and exploit the off-policy RL to learn the optimal intra-cluster CIO tuning policy. Specifically, the SBS load distribution and user distribution are defined as the state, the change of CIO values is defined as the action, and the load balancing utility is defined as the reward. At each time $ t $, the bottom layer selects the MLB action $ \ba_t $ according to the observed UDN state $ \bs_t $, which shifts the system to the next state $ \bs_{t+1} $ according to the state transition probability $ p(\bs_{t+1}|\bs_t, \ba_t) $. Note that the transition probability is conditioned on both the current UDN state $ \bs_t $ and the MLB action $ \ba_t $, which indicates that the MLB actions affect not only the immediate MLB reward but also the next UDN state and, through that, all subsequent MLB rewards.
	Compared with the traditional formulations which usually aim at maximizing the load balancing utility at a specific time slot, the RL-based formulation is more far-sighted, aiming at achieving a balanced load distribution over a long time horizon.
	Moreover, the RL-based formulation learns from the incrementally generated data samples when interacting with the UDN environment for MLB, thereby bearing a built-in data-driven learning nature.
	The detailed learning context is defined as follows.
	\subsubsection{States}
	The inputs of the RL-based learning algorithm are the wireless system status. 
	However, directly using the raw system data as the RL inputs is problematic: 
	first, the state space could be too large to enumerate;
	second, the high-dimensional raw data is computationally challenging to process and time costing to transmit;
	third, the raw data may contain too much redundancy to negatively impact the learning performance.
	Therefore, in this paper, we propose to use a collection of high-level features to represent the wireless system status as the RL inputs. 
	Specifically, the state for each SBS $ i \in \mathcal{I} $ is composed of: 
	i) the SBS load derived from the averaged load $ \tilde{\rho}_i = \rho_i - \rho_g $, where $ \rho_g = \frac{1}{N} \sum_{i=1}^{N} \rho_i $ with $ N $ denoting the number of SBS;
	ii) the fraction of the edge users $ E_i$.
	Here we categorize the edge users according to their downlink SINRs to the corresponding serving SBS and the neighboring SBSs. Generally, one SBS with more edge users indicates that its handover is more sensitive to the change of CIOs. 
	In summary, the state vector could be written as
	\begin{equation}\label{eq:RL_states}
	\bs_t = \left[ \tilde{\rho}^t_1, \tilde{\rho}^t_2,\cdots, \tilde{\rho}^t_N, E^t_1, E^t_2, \cdots, E^t_N \right] ^{\top}.
	\end{equation} 
	It is noteworthy to point out that the feature selection is not unique; other combinations could be explored in the future.
	
	\subsubsection{Actions}
	The output of the RL-based learning algorithm is the CIO values, i.e., 
	\begin{equation}\label{eq:RL_action}
	\ba_t = \{ O_{ij}(t) ~|~ \forall i,j \in \mathcal{I} \}.
	\end{equation} 
	Here we have $ O_{ij} \in \left[ O_{\min}, O_{\max} \right] , \forall i,j \in \mathcal{I} $, where $ O_{\min} $ and $ O_{\max} $ are the predefined lower bound and upper bound, respectively.
	Note that, in this paper, in order to study a general control problem, we consider the CIO value to be continuous.
	
	\subsubsection{Rewards}
	The load distribution could be balanced by minimizing the maximum load (or equivalently maximizing the inverse of the maximum load) of all the SBSs~\cite{8094955}, i.e., alleviating the worst case.
	Hence, we define the reward function to be 
	\begin{equation}\label{eq:reward}
	r(\bs_t, \ba_t) = \frac{1}{\max_{i \in \mathcal{I}} \rho^t_i}.
	\end{equation}
	
	\subsubsection{Policy}
	The policy in RL is usually modeled as a stochastic function $ \pi: \mathcal{S} \rightarrow \mathcal{P(\mathcal{A})} $, which characterizes the probability of selecting an action $ a_t \in \mathcal{A} $ at an arbitrary state $ s_t \in \mathcal{S} $. 
	However, in this paper, we consider a \textit{deterministic} policy for CIO tuning in practice, which can be parameterized as
	\begin{equation}\label{key}
	\pi_{\theta}: \mathcal{S} \rightarrow \mathcal{A},
	\end{equation}
	where $ \theta \in \mathbb{R}^{n} $ is the parameter to optimize. 
	Particularly, in this paper, the policy is parameterized by the deep neural network to improve the model generalization capability, which is discussed in Sec.~\ref{sec:fueled}.
	
	\subsubsection{Objective}
	Our goal is to find a CIO tuning policy which can maximize the cumulative discounted MLB reward from the start state $ s_0 $. 
	The objective function can be written as
	\begin{subequations} \label{eq:J_single}
		\begin{align}
		J_{\beta}(\pi_{\theta}) &= \int_{\mathcal{S}} \kappa^{\beta}(s)V^{\pi_{\theta}}(s) ds \\
		&= \int_{\mathcal{S}} \kappa^{\beta}(s) \int_{\mathcal{A}} \pi(s,a)Q^{\pi_{\theta}}(s,a) da ds \\
		&= \int_{\mathcal{S}} \kappa^{\beta}(s) Q^{\pi_{\theta}}\left( s, \pi_{\theta}(s) \right)  ds \\
		&= \mathbb{E}_{s \sim \kappa^{\beta}} \left[ \sum_{k=0}^{\infty} \gamma^{k}r\left( s, \pi_{\theta}(s) \right)  \right],
		\end{align}
	\end{subequations}
	where $ \beta: \mathcal{S} \rightarrow \mathcal{P(\mathcal{A})} $ is a stochastic behavior policy, and 
	\begin{equation}\label{key}
	\kappa^{\beta}(s) := \int_{\mathcal{S}} \sum_{t=1}^{\infty} \gamma^{t-1} p_1(s) p(s \rightarrow s', t, \beta) ds,
	\end{equation}
	is the discounted state visitation distribution~\cite{silver2014deterministic} when following the behavior policy $ \beta $ with  $ p_1(s) $ the probability of the starting state and
	$ p(s \rightarrow s', t, \beta) $ the transition probability from state $ s $ to state $ s' $ under policy $ \beta $ at time $ t $. 
	Generally, the objective can be viewed as the value function or the Q-function of the target policy averaged over the state distribution of the behavior policy~\cite{silver2014deterministic}.
	Finally, the off-policy RL-based intra-cluster MLB problem could be written as
	\begin{align}\label{P0}
	\mathcal{P}_{0}: \quad  &\max_{\theta} J_{\beta}(\pi_{\theta}) \\
	\mathrm{s.t.}  \quad & C_{1}:\mathcal{X}^t_{u, i} \in \lbrace 0,1 \rbrace , \sum_{i \in \mathcal{I}} \mathcal{X}^t_{u, i} \leq 1,  \forall u \in \mathcal{U},\\
	& C_{2}: O_{ij} \in [O_{\min}, O_{\max}], \forall i,j \in \mathcal{I},
	\end{align}
	where $ \mathcal{X}^t_{u, i} $ defines the uniqueness of the user-to-SBS association, $ C_{1} $ defines the uniqueness of user association, and $ C_{2} $ defines the bound over the CIO values.
	
	\section{Centralized Dynamic Load-Driven Clustering}
	\label{sec:top-layer}
	In our proposed architecture, the top layer aims at dynamically grouping all the SBSs into a bunch of clusters according to the global traffic variations.
	The clustering can be triggered either periodically or based on certain traffic patterns, e.g., the sudden increase of traffic due to some sport events, the daily tidal-effect, etc. 
	In this paper, we tailor the celebrated Lloyd's k-means clustering method~\cite{1056489} to perform \textit{load-driven} clustering for the underlying SBSs, which is fast to converge in practice.
	The proposed clustering algorithm consists of two phases: 
	1) load-based initialization, which is based on the detection of traffic hotspots, i.e., the top overloaded SBSs; 
	2) location-based SBS clustering, which aims at grouping the hotspot SBSs with their nearby SBSs based on the \textit{geographical} distance.
	
	\subsubsection{Load-based Initialization}
	The selection of initial cluster centroids can largely influence the clustering result. A commonly used strategy is to randomly choose a bunch of SBSs as the initial centroids~\cite{celebi2013comparative}. 
	Based on the domain knowledge, in this paper, we instead propose to choose the top $ k $ overloaded SBSs as the initial centroids in order to efficiently offload the traffic from the hotspot SBSs for load balancing.
	First, we define the \textit{stage-averaged load} to be the averaged SBS load over the previous MLB stage.
	Specifically, for each SBS $ i \in \mathcal{I} $, we define its stage-averaged load over the previous stage $ k-1 $ as
	\vspace*{-0.5\baselineskip}
	\begin{equation}\label{eq:averaged_load}
	\hat{\rho}^{k-1}_i  = \frac{1}{T_{k-1}} \sum_{t = t^0_{k-1}}^{t^0_{k-1}+T_{k-1}-1}\rho^t_i,
	\end{equation}
	where $ t^0_{k-1} $ is the start time of stage $ k-1 $, and $ T_{k-1} $ is the time length of stage $ k-1 $. 
	The stage-averaged load can reflect a fair load level by reducing the influence from abnormal traffic variations.
	Next, we rank the SBSs according to their stage-averaged loads to obtain a sorted list $ \boldsymbol{C}_{\text{list}} $ as the initial cluster centroids.
	
	\subsubsection{Location-based Clustering}
	Given the list of candidate centroids from the initialization phase, we next group the SBSs according to their geographical locations by minimizing the sum of squared error (SSE):
	\begin{align}
	&\min_{\mathcal{C}_h, \boldsymbol{c}_h}
	\sum_{h=1}^{H} \sum_{\boldsymbol{x}_{i} \in \mathcal{C}_h} \norm{\boldsymbol{x}_i - \boldsymbol{c}_h}_2^2 \\
	\mathrm{s.t.}  \quad & \mathcal{X}^t_{i, h} \in \lbrace 0,1 \rbrace , \sum_{h \in \mathcal{H}} \mathcal{X}^t_{i, h} \leq 1,  \forall i \in \mathcal{C}_h,
	\end{align}
	where $ H $ is the predetermined number of clusters, $ \mathcal{C}_h $ is the set of SBSs that are assigned to the cluster $ h $, $ \boldsymbol{x}_i $ is the location of SBS $ i $, $ \mathcal{X}^t_{i, h} $ defines the uniqueness of the SBS-to-cluster association, and
	\begin{equation}\label{key}
	\boldsymbol{c}_h = \frac{1}{|\mathcal{C}_h|} \sum_{\boldsymbol{x}_{i} \in \mathcal{C}_h} \boldsymbol{x}_i,
	\end{equation}
	is the centroid location of cluster $ \mathcal{C}_k $ with $ |\cdot| $ the cardinality.
	The detailed procedure is provided in Algorithm~\ref{algorithm:clustering}.
	
	Note that the selection of $ H $ is also critical.
	The clustering validation methods, e.g., the Calinski-Harabasz index~\cite{1114856}, can be employed to evaluate all the possible $ H $ and choose the best one. 
	For example, we can first set a searching range for the candidate $ H $, e.g., $ \mathcal{H} = \{1,2,\cdots, H_{\text{max}}\} $ where $ H_{\text{max}} $ is the upper bound; 
	then, we evaluate the Calinski-Harabasz index of the clustering result with each $ h \in \{ 1, 2, \cdots, H_{\text{max}}\}$; 
	finally, we choose the one with the highest Calinski-Harabasz index as the optimal choice. 
	
	It is noteworthy to remark that our proposed k-means algorithm is only one possible solution for the dynamic clustering at the top layer; other explorations could be done in the future with different preferences, e.g., by considering user mobility pattern, social relationships, radio propagation, etc.
	\begin{algorithm}
		\caption{ Load-driven Clustering Based on K-means }
		\label{algorithm:clustering}
		\begin{algorithmic}[1]
			\STATE \textbf{Input:} 
			\STATE { $ t $; $ \rho^{t}_i $; $ \boldsymbol{x}_i $; number of clusters $ H $; number of SBSs $ N $.}
			\STATE \textbf{Initialization:} 
			\STATE { Calculate the stage-averaged loads according to (\ref{eq:averaged_load}).}
			\STATE { Rank the stage-averaged loads to obtain $ \boldsymbol{C}_{\text{list}} $.}
			\STATE { Initialize the cluster centroids with the top $ H $ elements from $ \boldsymbol{C}_{\text{list}} $.  }
			\STATE \textbf{Iteration:} 
			\REPEAT
			\FOR {$ i = 1,2,\cdots,N $}
			\FOR {$ h = 1,2,\cdots,H $}
			\STATE { Update the SBS-to-cluster association by \[ \mathcal{X}^t_{i, h} = \argmin \norm{ \boldsymbol{x}_i - \boldsymbol{c}_h }^2_2 \]}
			\ENDFOR
			\ENDFOR
			\FOR {$ h = 1,2,\cdots,H $}
			\STATE {Update the cluster centroids by \[ \boldsymbol{c}_h = \frac{\sum_{i=1}^{N}\mathcal{X}^t_{i, h} \boldsymbol{x}_i}{\sum_{i=1}^{N}\mathcal{X}^t_{i, h}} \]}
			\ENDFOR
			\UNTIL No change of the cluster centroids.
		\end{algorithmic}
	\end{algorithm}
	
	% ===========================
	%          section
	% ===========================
	\section{Self-Organized DRL-Based Load Balancing}
	\label{sec:bottom-layer}
	The bottom layer aims at balancing the load within each cluster with the proposed off-policy DRL-based MLB algorithm. 
	Generally, the classic off-policy method optimizes the target policy by following a \textit{single} behavior policy for exploration. However, it is typical to assume that the behavior policy must allow the RL agent to explore every possible state infinitely often~\cite{sutton2018reinforcement}. To this end, the commonly adopted behavior policy should be a fully randomized one, such that the explored high-quality samples are usually very sparse, which makes the learning inefficient. 
	Therefore, in this paper, we propose to adopt \textit{multiple} behavior policies for joint exploration, which covers the use of a single behavior policy as a special case, and has the following advantages.
	\begin{enumerate}
		\item \textbf{Stability:} The learning now only requires the jointly explored space of multiple behavior policies to cover the entire state-space. Hence, the learning over multiple behavior policies can perform well even though none of the behavior policies can lead to the optimal solution on its own. 
		\item \textbf{Efficiency:} The explored samples by following different behavior policies are richer than simply following one behavior policy, which implies a higher probability to find high-quality samples. 
		Moreover, the behavior policies could include traditional MLB methods to directly generate high-quality samples to realize \textit{expert-guided} or \textit{model-assisted} learning, which could improve the learning efficiency.
	\end{enumerate}
	Note that the learning under multiple behavior policies can be implemented for practical systems in many ways. 
	For example, 
	1) the self-organized cluster could use different behavior policies in turn, e.g., using traditional MLB methods to guide learning at the beginning and switching to random exploration later;
	2) different self-organized clusters under similar MDPs, e.g., due to similar cell layouts and user mobility patterns, can learn jointly, where each of them follows a different behavior policy and updates a shared target policy.\footnote{In this case, the learning under multiple behavior policies can be considered as a simple multi-agent learning paradigm.};
	3) the target policy can be trained offline, e.g., in a simulation platform, by following multiple behavior policies, and then be used for online control.
	
	In the sequel, we first extend the policy gradient theorem for the case with multiple behavior policies, which is the key for policy improvement; then we present a parallel learning framework. Finally, we empower the learning framework with deep neural networks to largely improve its generalization capability in complex UDN environments.
	
	\subsection{Policy Gradient with Multiple Behavior Policies}
	\subsubsection{Preliminaries on Policy Gradient}
	Traditional policy improvement methods find a better policy by greedily choosing the best action under all possible states according to the estimated value function or Q-function \cite{sutton2018reinforcement}. However, such a greedy searching strategy becomes computationally demanding when the action space is large, and usually impossible for a continuous action space. 
	Therefore, in this paper, we exploit the \textit{policy gradient method} for policy improvement, which overcomes the limitations of the greedy searching strategy by explicitly optimizing a parameterized policy. 
	Specifically, the policy gradient theorem~\cite{sutton2018reinforcement} gives the analytic expression for the gradient of the objective $ J(\pi_{\btheta}) $ with respect to the policy parameters $ \btheta $, written as
	\begin{equation}\label{eq:policy_gradient}
	\nabla_{\btheta} J(\pi_{\btheta}) = \mathbb{E}_{\bs \sim d^{\pi_{\btheta}}, \ba \sim \pi_{\btheta}} [\nabla_{\btheta} \log \pi_{\btheta} (\ba|\bs) Q^{\pi_{\btheta}}(\bs, \ba)],
	\end{equation}
	where $ d^{\pi_{\btheta}} $ is the state distribution when following policy $ \pi_{\btheta} $.
	Later, Silver \textit{et al.} proposed the off-policy deterministic policy gradient (OPDPG) theorem~\cite{silver2014deterministic} to optimize a deterministic target policy by following a \textit{single} stochastic behavior policy, written as
	\begin{subequations}\label{OPDAC}
		\begin{align}
		\nabla_{\theta} J_{\beta}(\pi_{\theta}) &\approx \int_{\mathcal{S}} \rho^{\beta}(s) \nabla_{\theta} \pi_{\theta} (s) \nabla_{a} Q^{\pi}(s,a)|_{a=\pi_{\theta}(s)} \text{d}s  \\
		&= \mathbb{E}_{s \sim \rho^{\beta}} [\nabla_{\theta} \pi_{\theta}(s) \nabla_{a} Q^{\pi}(s,a)|_{a=\pi_{\theta}(s)}],
		\end{align}
	\end{subequations}
	where the approximation comes from the drop of one term depending on the Q-value gradient $ \nabla_{\theta}Q^{\pi}(s,a) $ to preserve the convergence~\cite{silver2014deterministic}. 
	
	\subsection{OPDPG with Multiple Behavior Policies}
	The OPDPG theorem can be adopted to optimize our objective function with a proper extension on the learning over multiple behavior policies.
	Specifically, we denote the set of multiple behavior policies to follow as $ \mathcal{M} = \{ \beta_1, \beta_2, \cdots, \beta_M \} $, where $ M $ is the total number of behavior policies. Based on the objective under a single behavior policy given in (\ref{eq:J_single}), the objective under multiple behavior policies could be written as
	\begin{equation}
	J(\pi_{\theta}) = \sum_{m \in \mathcal{M}} J_{\beta_m}(\pi_{\theta}),
	\end{equation}
	where 
	\vspace*{-0.5\baselineskip}
	\begin{equation} \label{eq:J_single2}
	J_{\beta_m}(\pi_{\theta}) = \mathbb{E}_{s \sim \kappa^{\beta_m}} \left[ \sum_{k=0}^{\infty} \gamma^{k}r\left( s, \pi_{\theta}(s) \right)  \right],
	\end{equation}
	with $ \kappa^{\beta_m}(s) := \int_{\mathcal{S}} \sum_{t=1}^{\infty} \gamma^{t-1} p_1(s) p(s \rightarrow s', t, \beta) ds $.
	The new objective $ J(\pi_{\theta})  $ can be viewed as the value function of the target policy averaged over the states generated by following multiple behavior policies.
	The intra-cluster MLB problem now becomes
	\begin{equation}\label{P1}
	\begin{aligned}
	\mathcal{P}_{1}: \quad  &\max_{\theta} J(\pi_{\theta}) = \sum_{m \in \mathcal{M}} J_{\beta_m}(\pi_{\theta}) \\
	\mathrm{s.t.}  \quad & C_{1}: \mathcal{X}^t_{u, i} \in \lbrace 0,1 \rbrace , \sum_{i \in \mathcal{I}} \mathcal{X}^t_{u, i} \leq 1,  \forall u \in \mathcal{U},\\
	& C_{2}: O_{ij} \in [O_{\min}, O_{\max}], \forall i,j \in \mathcal{I}.
	\end{aligned}
	\end{equation}
	Accordingly, the policy gradient for the learning under multiple behavior policies is given as
	\begin{subequations}\label{eq:extended_OPDPG}
		\begin{align}\label{}
		& \nabla_{\theta} J(\pi_{\theta}) =  \sum_{m \in \mathcal{M}} \nabla_{\theta} J_{\beta_m}(\pi_{\theta}) \\
		&\approx \sum_{m \in \mathcal{M}} \int_{\mathcal{S}} \rho^{\beta_m}(s) \nabla_{\theta} \pi_{\theta} (s) \nabla_{a} Q^{\pi}(s,a)|_{a=\pi_{\theta}(s)} \text{d}s \label{eq:aa}\\
		&= \sum_{m \in \mathcal{M}} \mathbb{E}_{s \sim \rho^{\beta_m}} [\nabla_{\theta} \pi_{\theta}(s) \nabla_{a} Q^{\pi}(s,a)|_{a=\pi_{\theta}(s)}], 
		\end{align}
	\end{subequations}
	where in (\ref{eq:aa}), we apply the OPDPG theorem for each behavior policy, respectively.
	\begin{figure*}[tb]
		\centering
		\includegraphics[trim = 10 10 10 10, clip, width=16cm]{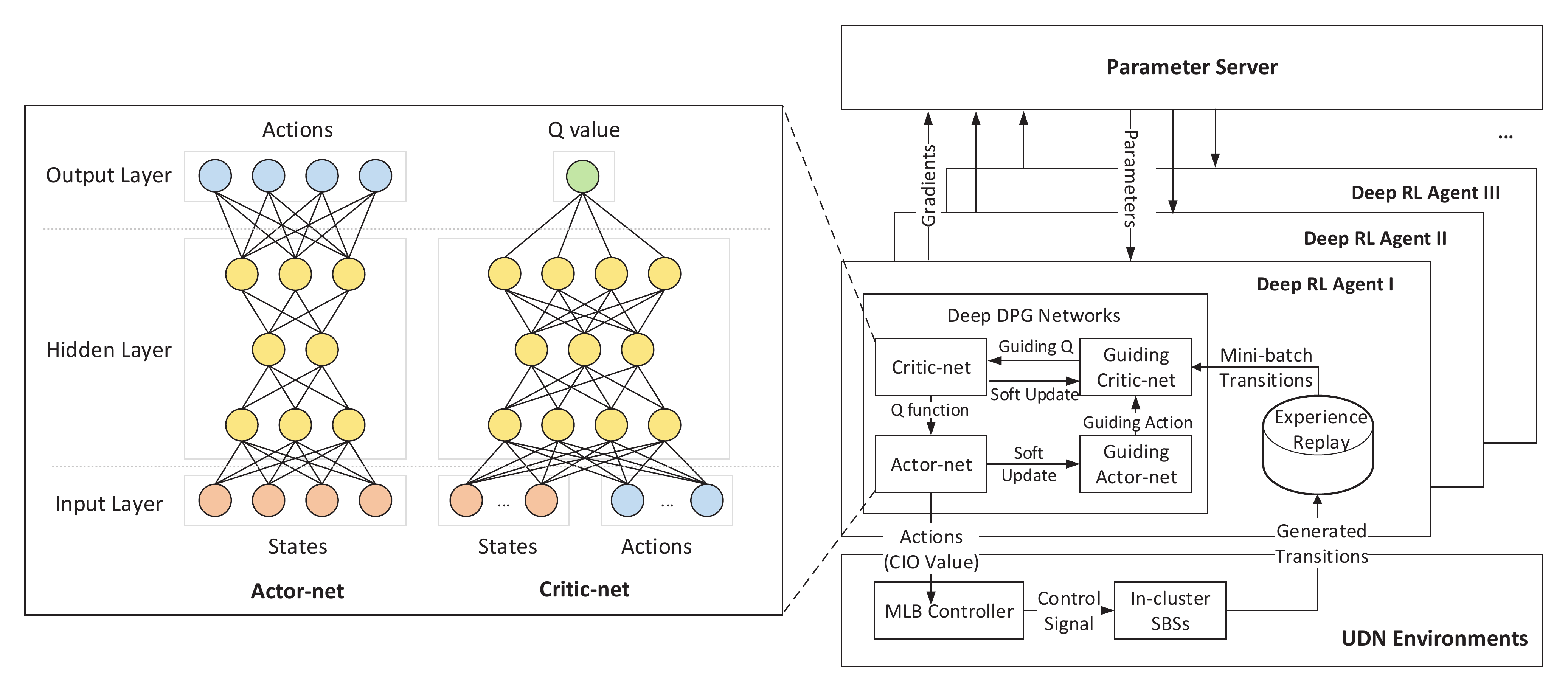}
		\caption{Overview of the DRL architecture for MLB}
		\label{fig:structure}
	\end{figure*}
	
	\begin{figure*}[t]
		\centering
		\subfigure[Actor-Critic]
		{ 	
			\label{fig:actor-critic}
			\includegraphics[trim = 3 3 3 5, clip, width=5cm]{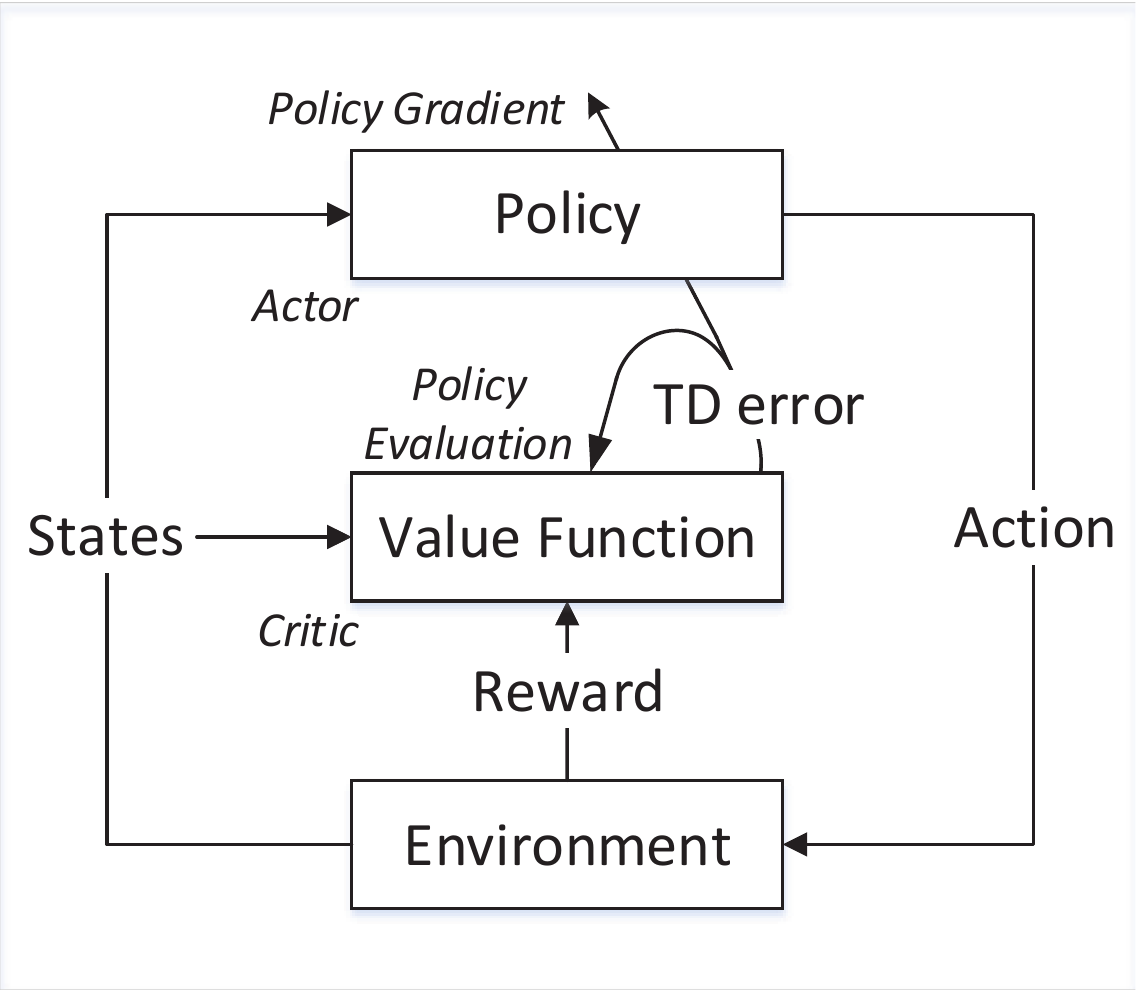}
		}\hspace{5mm}
		\subfigure[Explore With Multiple Behavior Policies]
		{ 	
			\label{fig:parallelAC}
			\includegraphics[trim = 10 2 10 18, clip, width=10cm]{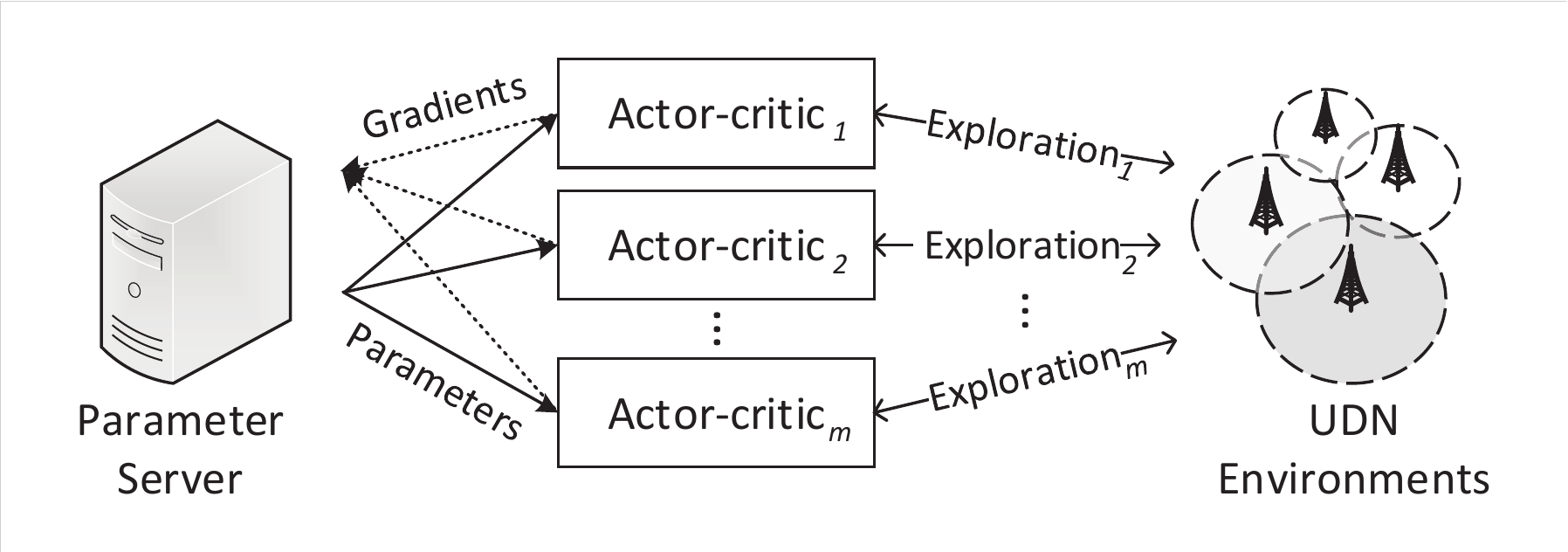}
		}
		\caption{Parallel learning framework}
		\label{fig:parallel-RL}
	\end{figure*}
	\subsection{Parallel Learning Framework}
	We now develop a parallel learning framework based on the actor-critic algorithm \cite{sutton2018reinforcement}, where each agent follows a different behavior policy for exploration, while updating a shared target policy simultaneously.
	\subsubsection{Preliminaries on Actor-Critic}
	The actor-critic algorithm~\cite{sutton2018reinforcement}, which comprises an actor and a critic learner, is one of the most popular learning algorithms based on the policy gradient methods.
	As shown in Fig.~\ref{fig:actor-critic}, the critic estimates the true Q-function $ Q^{\pi_{\btheta}}(\bs, \ba) $ of the actor's policy $ \pi_{\btheta} $ with a parameterized function $ Q^{\bw}(\bs, \ba) $, where $ \bw \in \mathbb{R}^n $ is the parameters to be learned based on the policy evaluation methods~\cite{sutton2018reinforcement}. The actor, on the other hand, improves its policy $ \pi_{\btheta} $ with the estimated Q-function $ Q^{\bw}(s,a) $ based on the policy gradient methods.
	In this paper, we use Q-learning for policy evaluation in the critic, while using the extended OPDPG in (\ref{eq:extended_OPDPG}) for the policy gradient in the actor.
	The critic and the actor will iteratively improve the estimated Q-function $ Q^{\bw}(\bs, \ba) $ and the policy $ \pi_{\btheta} $ until convergence.
	
	\subsubsection{Parallel Learning Framework with Multiple Behavior Policies}
	As shown in Fig.~\ref{fig:parallelAC}, we employ multiple actor-critic RL agents to learn in parallel. Generally, each RL agent interacts with the environment by using a different behavior policy and transmits the calculated gradients to a centralized parameter server. The parameter server uses the local gradients to update a set of global parameters and periodically synchronize with the local parameters for consensus. The above procedures are performed repeatedly until convergence or some termination criteria are met. In particular, the detailed work-flow for the iteration at time $ t $ is presented as follows. 
	
	First, for each agent $ m \in \mathcal{M} $ at time $ t $, the local actor observes the current state $ s_t $, and executes an action $ a^m_t $ by following its own behavior policy with $ a^m_t = \beta_m(s^m_t) $. Afterwards, the current state $ s^m_t $ shifts to $ s^m_{t+1} $ and returns a reward $ r^m_t $. 
	The generated transition at time $ t $ can be written as $ (s^m_t, a^m_t, s^m_{t+1}, r^m_t) $.
	
	Second, each local critic evaluates the Q-function of the target policy based on Q-learning~\cite{sutton2018reinforcement}. Specifically, the gradient $ \Delta^m_{w} $ for the Q-function updates can be written as
	\vspace*{-0.5\baselineskip}
	\begin{align}
	\delta^m_t &\!=\! r^m_t \!+\! \gamma Q^{w} (s^m_{t+1}, \pi_{\theta}(s^m_{t+1})) \!-\! Q^{w}(s^m_t, a^m_t) \label{OPDAC:critic1}, \\
	\Delta^m_{w} &= \delta^m_t \nabla_{w} Q^{w}(s^m_t, a^m_t) \label{OPDAC:critic2},
	\end{align}
	where $ \delta^m_t $ is the temporal difference (TD) error, with $ \pi_{\theta}(s^m_{t+1}) $ the action selected by the target policy $ \pi_{\theta} $ at state $ s^m_{t+1} $. Note that each local critic only needs to submit the gradient to the parameter server, without applying to the local Q-function.
	
	Third, each local actor computes the policy gradient according to the extended OPDPG theorem in (\ref{eq:extended_OPDPG}). The gradient $ \Delta^m_{\theta} $ for the target policy updates can be written as
	\vspace*{-0.5\baselineskip}
	\begin{equation}\label{eq:gradient-actor}
	\Delta^m_{\theta} = \nabla_{\theta} \pi_{\theta} (s^m_t) \nabla_a Q^{w}(s^m_t, a)|_{a=\pi_{\theta}(s^m_t)}.
	\end{equation}
	Similarly, each local actor only needs to submit the gradient to the parameter server, without applying to the local policy.
	
	Finally, the parameter server uses the collected gradients to update a set of global parameters as
	\vspace*{-0.5\baselineskip}
	\begin{align}
	w^g_{t+1} &= w^g_{t} + \sum_{m=1}^{M} \alpha_w \Delta^m_w, \label{eq:update-global-w} \\
	\theta^g_{t+1} &= \theta^g_{t} + \sum_{m=1}^{M} \alpha_{\theta} \Delta^m_{\theta}, \label{eq:update-global-theta}
	\end{align}
	where $ \alpha_w $ and $ \alpha_{\theta} $ are the step-sizes.
	Afterwards, the local parameters $ w^m_t $ and $ \theta^m_t $ of all the agents $ m \in \mathcal{M} $ (including all the actors and critics) are synchronized with the global parameters $ w^g_{t+1} $ and $ \theta^g_{t+1} $ for consensus, which completes this iteration.
	
	\subsubsection{Convergence Analysis}
	Suppose that $ p(s'|s,a) $, $ \nabla_a p (s'|s,a) $, $ \mu_{\theta} $, $ \nabla_{\theta} \mu_{\theta}(s)$, $ r(s,a) $, $ \nabla_{a} r(s,a) $, $ p_1(s) $ are continuous in variables $s, a$ and $s'$, and that the state space $ \mathcal{S} $ is a compact subset of $ \mathbb{R}^d $, the convergence of the proposed parallel actor-critic learning framework can be reached when we use: i) compatible linear Q-function approximators~\cite{silver2014deterministic}; ii) the gradient temporal-difference (GTD) based learning method~\cite{6949624} for the critic updates.
	
	The justification follows the one in~\cite{silver2014deterministic}.
	First, the use of a compatible linear Q-function approximator ensures that the deterministic policy gradient will not be affected when using the estimated gradient $ \nabla_a Q^w(s,a) $ to replace the true gradient $ \nabla_a Q^{\mu}(s,a) $. 
	In particular, a function approximator that is compatible with the deterministic policy $ \mu_{\theta}(s) $ is defined to satisfy~\cite{silver2014deterministic}: 
	\begin{enumerate}
		\item $ \nabla_a Q^w(s,a)|_{a=\mu_{\theta}(s)} \!=\! \nabla_{\theta} \mu_{\theta}(s)^{\top} w $; 
		\item  $ w $ minimizes the mean-square error $ \text{MSE}(\theta, w) = \mathbb{E}_{s \sim \rho^{\beta_m}}[\epsilon(s;\theta,w)^{\top}\epsilon(s;\theta,w)] $, where $ \epsilon(s;\theta,w) \!=\! \nabla_a Q^w(s,a)|_{a=\mu_{\theta}(s)} \!-\! \nabla_a Q^{\mu}(s,a)|_{a=\mu_{\theta}(s)} $.
	\end{enumerate}
	Therefore, we have 
	\begin{equation}\label{key}
	\nabla_{w} \epsilon(s;\theta,w) = \nabla_{w} \nabla_a Q^w(s,a)|_{a=\mu_{\theta}(s)} = \nabla_{\theta} \mu_{\theta}(s).
	\end{equation}
	If $ w $ can minimize the MSE, the gradient of MSE \textit{w.r.t.} $ w $ should satisfy
	\begin{subequations}
		\begin{align}\label{key}
		\nabla_{w} \text{MSE}(\theta, w) &= 2 \mathbb{E}_{s \sim \rho^{\beta_m}}[\nabla_{w} \epsilon(s;\theta,w) \epsilon(s;\theta,w)] \\ 
		&= 2\mathbb{E}_{s \sim \rho^{\beta_m}}[\nabla_{\theta} \mu_{\theta}(s)\epsilon(s;\theta,w)] = 0.
		\end{align}
	\end{subequations}
	According to the definition of $ \epsilon(s;\theta,w) $, we have
	\begin{subequations}
		\begin{align}\label{key}
		&\mathbb{E}_{s \sim \rho^{\beta_m}}[\nabla_{\theta}\mu_{\theta}(s) \nabla_a Q^w(s,a)|_{a=\mu_{\theta}(s)}] \\
		= &\mathbb{E}_{s \sim \rho^{\beta_m}}[\nabla_{\theta}\mu_{\theta}(s) \nabla_a Q^{\mu}(s,a)|_{a=\mu_{\theta}(s)} ] \\
		= &\nabla_{\theta}J_{\beta^m}(\mu_{\theta}),	
		\end{align}
	\end{subequations}
	which proves that the estimated gradient $ \nabla_a Q^w(s,a) $ can replace the true gradient $ \nabla_a Q^{\mu}(s,a) $ without affecting the deterministic policy gradient. 
	Note that the above proof applies to any of the multiple behavior policies $ \beta_m \in \mathcal{M} $.
	
	Second, the GTD based policy evaluation has been proven to have sure convergence for the case with multiple behavior policies when using true gradients~\cite{6949624}, which herein is also applicable when using estimated gradients based on the above proof.
	Hence, we can use the extended OPDPG theorem in (\ref{eq:extended_OPDPG}) for the policy gradient, and use the diffusion off-policy GTD in~\cite{6949624} for the policy evaluation to ensure the convergence. 
	Note that, according to~\cite{6949624}, the actor and the critic should be updated with a sufficiently small step-size, in order to minimize the mean-squared projected Bellman error (MSPBE).
	
	In order to achieve a better generalization capability, in the following part, we further apply the nonlinear deep neural networks as the function approximators; and use Q-learning~\cite{sutton2018reinforcement} for critic updates, instead of the diffusion off-policy GTD~\cite{6949624}, to reduce the training complexity.
	Consequently, the convergence is no longer theoretically guaranteed, which is a common issue for deep learning based models. But the empirical results show that our proposed parallel actor-critic learning framework with deep neural networks can still converge in practice.   
	
	\subsection{Fueled With Deep Neural Networks}
	\label{sec:fueled}
	\begin{algorithm}
		\caption{DRL-based MLB With Parallel Learning Framework}
		\label{RL_algorithm}
		\begin{algorithmic}[1]
			\STATE \textbf{Input:} 
			\STATE {for $ \forall m \in \mathcal{M}$, $ \alpha^m_w =10^{-3} $, $ \alpha^m_{\beta} =10^{-4} $; $ \gamma = 0.99 $; $ \tau = 0.001 $; $ K = 64 $.}
			\STATE \textbf{Initialization:} 
			\STATE {Random initialize the weights $ w $ and $ \theta $; for $ \forall m \in \mathcal{M} $, set $ w_m \leftarrow w , \theta_m \leftarrow \theta $, $ \hat{w}_m \leftarrow w_m,  \hat{\theta}_m \leftarrow \theta_m$.  }
			\STATE \textbf{Iteration:} 
			\STATE {Observe the start state $ s^m_1$.}
			\FOR {$t = 1,  \infty $}
			\FOR {$ m = 1, M $}
			\STATE {Synchronize $ w_m \leftarrow w , \theta_m \leftarrow \theta $.}
			\STATE {Update the target networks with (\ref{eq:update-target-w}) and (\ref{eq:update-target-theta}).}
			\STATE {Observe $ s^m_t $ and execute $ a^m_t = \beta_m(a_t|s_t)$.}
			\STATE {Receive $ r^m_t$ and shift to the next state $ s^m_{t+1} $.}
			\STATE {Store the transition $ (s^m_t, a^m_t, r^m_t, s^m_{t+1}) $ in $ \mathcal{D}_m $.}
			\STATE {Randomly select $ K $ transitions from $ \mathcal{D}_m $.}
			\STATE {Calculate the local $ \Delta^m_{w} $ for critic with (\ref{eq:batch-critic-gradient}).}
			\STATE {Calculate the local $ \Delta^m_{\theta} $ for actor with (\ref{eq:batch-actor-gradient}).}
			\STATE {Send $ \Delta^m_{w} $ and $ \Delta^m_{\theta} $ to the parameter server.}
			\ENDFOR
			\STATE {Update the global $ w $ and $ \theta $ with (\ref{eq:update-global-w}) and (\ref{eq:update-global-theta}).}
			\ENDFOR
		\end{algorithmic}
	\end{algorithm}
	
	Theoretical convergence and sample-complexity analysis in RL have been mainly developed when using tabular or linear functions to approximate the Q-functions~\cite{sutton2018reinforcement}. The use of non-linear function approximators, e.g., neural networks, has long been considered to be unstable or even leading to divergence in RL due to the strong correlations among the generated samples~\cite{mnih2015human, Lillicrap16, mniha16, nair2015massively}. 
	However, recent successes on DRL have proven that using deep neural networks as function approximators is feasible when adopting the following two learning strategies~\cite{mnih2015human}:
	1) using mini-batch learning along with an experience replay, which can randomize the generated data samples and smooth the changes over data distributions;
	2) using guiding Q-networks, which can reduce the correlation between the learning Q-value $ Q(s,a) $ and the guiding Q-value $ r(s_t, a_t ) + \gamma Q(s_{t+1} , a_{t+1})$, when deriving the gradients. 
	In the sequel, we empower the proposed algorithm with the powerful deep neural networks, which are trained based on the aforementioned two learning strategies.
	
	\subsubsection{Deep Architecture}
	We use two deep neural networks to respectively represent the policy function in the actor and the Q-function in the critic to improve the generalization capability for MLB, which are referred to as the \textit{actor-net} and \textit{critic-net}, respectively.
	The high-level structure is presented in Fig.~\ref{fig:structure}. Specifically, for the actor-net, the policy network takes the system states (the high-level features defined in Sec.~\ref{sec:problem_formulation}) as the inputs, and outputs the CIOs. For the critic-net, the Q-network takes both the system states and the output CIOs from the policy network as the inputs, and generates a scalar Q-value as the final output. The neurons are fully connected. 
	\begin{figure*}[tb]
		\centering
		\includegraphics[trim = 10 10 10 10, clip, width=16.5cm]{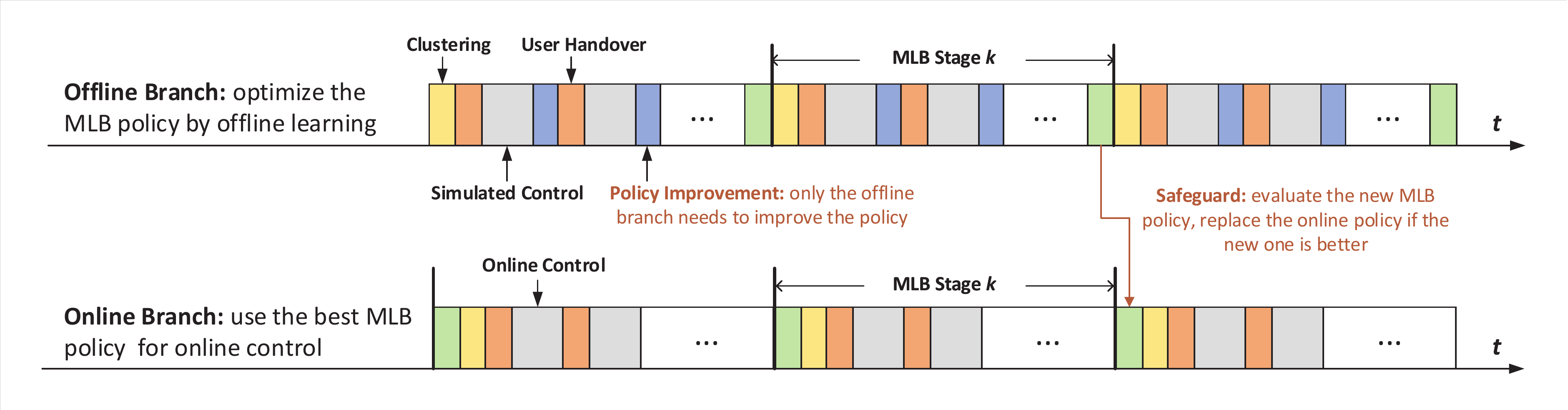}
		\caption{Safeguard mechanism for online control}
		\label{fig:safeguard}
	\end{figure*}
	\subsubsection{Guiding Networks}
	We use the guiding networks to stabilize the learning process in the actor and the critic. 
	Specifically, we introduce two more deep neural networks with the same structure as the critic-net and actor-net, respectively, to be the \textit{guiding critic-net} $ Q^{\hat{w}_m}(s,a)$ and the \textit{guiding actor-net} $ \pi_{\hat{\theta}_m}(s) $.
	The aim of the guiding network is to provide a stabilized learning target when updating the critics. 
	In particular, according to the critic update rules in (\ref{OPDAC:critic1}) and (\ref{OPDAC:critic2}), the loss function to minimize is given by
	\begin{equation}\label{eq:loss-function}
	\mathcal{L}(w_m) \!=\! \mathbb{E}_{(s_i, a_i, r_i, s_{i+1}) \in \mathcal{D}_m}\!\!\left[ \left( y^m_i - Q^{w_m}(s^m_i, a^m_i) \right)^2 \right],
	\end{equation}
	where
	\vspace*{-0.5\baselineskip}
	\begin{equation}\label{key}
	y^m_i = r(s^m_i, a^m_i) + \gamma Q^{\hat{w}_m}(s^m_{i+1}, \pi_{\hat{\theta}_m}(s^m_{i+1})).
	\end{equation}
	Here $ y^m_i $ serves as the learning target for the critic-net. However, different from (\ref{OPDAC:critic1}) and (\ref{OPDAC:critic2}), the estimated Q-value in $ y^m_i $ is now generated by the guiding critic-net, with the action predicted by the guiding actor-net, i.e., $ Q^{\hat{w}_m}(s^m_{i+1}, \pi_{\hat{\theta}_m}(s^m_{i+1})) $. 
	The parameters of the guiding networks should slowly track the parameters of the learning networks, such that $ y^m_i $ becomes a stabilized learning target for the critic-net. In this way, the training of the critic-net is closer to a supervised learning problem, where a robust solution likely exists.
	In practice, the guiding parameters $ \hat{w}_m $ and $ \hat{\theta}_m $ can be updated as
	\begin{align} 
	\hat{w}^{t}_m &= \tau w^{t}_m + (1-\tau) \hat{w}^{t}_m, \label{eq:update-target-w}\\
	\hat{\theta}^{t}_m &= \tau \theta^{t}_m + (1-\tau) \hat{\theta}^{t}_m, \label{eq:update-target-theta}
	\end{align}
	where $ \tau \ll 1 $ is a fixed step-size.
	
	\subsubsection{Mini-batch Learning}
	We use the mini-batch gradient descent to train the deep neural networks, which is a widely used technique in deep learning.
	Generally, in mini-batch learning, the training dataset are first split into multiple mini-batches of certain size. The gradients over one mini-batch are summed or averaged to reduce the variance of the gradient. In this way, the mini-batch gradient descent is promising to find a balance between the efficiency of stochastic gradient descent and the robustness of full-batch gradient descent.
	In our model, each local agent $ m \in \mathcal{M} $ has a local experience replay $ \mathcal{D}^m $ to store the generated transitions $ (s^m_t, a^m_t, r^m_t, s^m_{t+1}) $ incrementally.
	At each time $ t $, for each local agent $ m \in \mathcal{M} $, we select $ K $ transitions randomly from its local experience replay to form one mini-batch. The gradients to minimize the loss function $ \mathcal{L}(w_m) $ are averaged as
	\vspace*{-0.5\baselineskip}
	\begin{equation}\label{eq:batch-critic-gradient}
	\Delta^m_w \!=\! \frac{1}{K}\!\! \sum_{i=1}^{K}\left[ \left( y^m_i \!-\! Q^{w_m}(s^m_i,\! a^m_i) \right) \!\nabla_{w_m} \!Q^{w_m}(s^m_i, a^m_i) \right]\!.
	\end{equation}
	Similarly, the gradients to update the target policy are averaged as 
	\begin{equation}\label{eq:batch-actor-gradient}
	\Delta^m_{\theta} \!=\! \frac{1}{K} \!\sum_{i=1}^{K} \alpha_{\theta} \nabla_{\theta_m} \pi_{\theta_m} (s^m_i) \nabla_a Q^{w_m}(s^m_i\!, a)|_{a=\pi_{\theta}(s^m_i)}.
	\end{equation}
	
	\subsubsection{Asynchronous Update}
	Based on the proposed parallel actor-critic learning framework, we use a parameter server to collect the mini-batch gradients $ \Delta^m_{w} $ and $ \Delta^m_{\theta} $ from the local agents to update a set of global parameters, and synchronize the local $ w $ and $ \theta $ of each agent with the global one for consensus. 
	Note that the training of the deep neural networks can be performed in an asynchronous manner~\cite{mniha16, nair2015massively}, which relaxes the coordination requirements.
	However, the received gradients exceeding a preset maximum time delay should be filtered out to ensure the learning stability~\cite{mniha16, nair2015massively}.
	
	The complete procedure of our proposed DRL-based intra-cluster MLB is presented in Algorithm~\ref{RL_algorithm}.
	
	% ===========================
	%          section
	% ===========================
	%\clearpage
	\section{Offline-Evaluation Based Online Control}
	\label{sec:safeguard}
	One of the major drawbacks of using the data-driven machine learning based method for online control is that the system needs to take the risk of applying under-trained policies for control at the early learning stage, which is unstable and even dangerous to the network, e.g., following an under-trained MLB policy may lead to heavily overloaded situations due to random handovers. 
	On the other hand, training the neural networks to re-adapt to the underlying UDN environments after each re-clustering process requires a certain amount of time, which may disturb the online MLB performance.
	Therefore, in this section, we propose an offline-evaluation based safeguard mechanism which runs an online branch for system control and an offline branch for policy learning in parallel. The online branch always uses the best MLB policy found and well-trained by the offline branch. 
	As such, the mechanism not only stabilizes the online performance, but also enables the exploration beyond-current or even over unexpected policies in a safe way.
	The philosophy behind is of vital importance to make full use of data-driven machine learning, which goes beyond existing methods to deliver superior alternatives.
	
	Specifically, we run the proposed MLB algorithm over the offline learning branch and the online controlling branch in parallel. The learning branch runs in an offline manner, e.g., by using a network evaluation system based on historical performance records; the controlling branch runs in the online system, which only uses the best MLB policy found by the offline branch. 
	For example, as shown in Fig.~(\ref{fig:safeguard}), the work-flow at stage $ k $ can be interpreted as follows.
	First, at the beginning of stage $ k $, the offline learning branch triggers the re-clustering process, and trains the neural networks to adapt to the new clustering result. 
	Then, at the end of stage $ k $, the safeguard evaluates the performance of the newly learned MLB policy over the previous stage $ k $. 
	If the newly learned policy can generate a better performance, the safeguard would replace the online policy with the new policy; otherwise, 
	it keeps the current online policy in the online branch and continues to seek a better one in the offline branch at the next stage.
	In this way, we can keep exploring better MLB policies in the offline branch, while only executing the best and well-trained policy online, thus ensuring the online system to operate with a stable MLB policy.
	% ===========================
	%          section
	% ===========================
	%\clearpage
	\section{Simulation Results}
	\label{sec:results}
	In this section, we evaluate the performance of the proposed two-layer MLB architecture along with the DRL-based MLB algorithm through simulations. 
	We choose four baseline methods: 1) the static rule-based algorithm~\cite{5594565}, which uses a fixed step-size to tune the CIOs; 2) the adaptive rule-based algorithm~\cite{6398873}, which uses an adaptive step-size to tune CIOs; 3) the Q-learning based algorithm~\cite{7393587}, which uses Q-learning to tune the CIOs between the most overloaded SBS and its neighboring SBSs; 4) a plain baseline without any MLB controls. We refer to them as the \textit{rule-based (static)} algorithm, the \textit{rule-based (adaptive)} algorithm, the \textit{Q-learning} algorithm, and the \textit{noMLB} algorithm, respectively. 
	
	The simulated SON scenario consists of $ 12 $ SBSs \textit{randomly} distributed in a $ 300m \times 300m $ area with $ 200 $ users randomly walking at $ 1~m/s\sim10~m/s $, each incurring a CBR traffic demand. 
	The transmit power of each SBS is set to be $ 46~dBm $. The path loss from each particular user to one particular SBS is modeled as $ 128.1 + 37.6 \log(\max\left\lbrace d, 0.035 \right\rbrace ) $, where $ d $ is the distance in $ km $. The extra log-normal shadowing is modeled with a zero mean and a standard deviation of $ 8~dB $. The handover hysteresis is set to be $ 3~dB $. 
	Both of the actor-net and critic-net are composed with two hidden layers, each with $ 400 $ and $ 300 $ neurons, respectively. All hidden layers are followed by a rectified non-linearity layer.
	The learning rate for actor and critic networks are fixed as $ 10^{-4} $ and $ 10^{-3} $, respectively. The discount factor $ \gamma $ for rewards is set to be $ 0.99 $. The $ \tau $ value for the soft target network update is set to be $ 0.001 $.
	The size of the mini-batch is set to be $ 64 $. The size of each local experience replay is set to be $ 10^5 $. 
	The simulations are performed on a workstation with eight Intel Xeon E3 CPU cores at 3.50 GHz and NVIDA GeForce GTX 970 with 4G graphics memory, Generally, the training of one DRL model with $ 10,000 $ time steps can usually be completed within 30 minutes.
	
	We consider three MLB performance indicators: 1) the reward function value; 2) the load standard deviation; 3) the handover failure ratio (HFR). Specifically, the HFR is obtained by introducing an admission control mechanism~\cite{5594565, 6398873} to block the incoming handover attempts if the SBS load exceeds 80\%, concretely, 
	\vspace*{-0.5\baselineskip}
	\begin{equation}\label{key}
	F = \frac{N_{\text{HOfail}}}{N_{\text{HOfail}} + N_{\text{HOsuccess}}},
	\end{equation}
	where $ N_{\text{HOfail}} $ is the number of blocked handover attempts of all the SBSs and $ N_{\text{HOsuccess}} $ is the number of successful handover attempts of all the SBSs.
	Moreover, the presented rewards are moving averaged to smooth out short-term fluctuations and highlight longer-term trends, thus presenting a clear and sharp comparison. Specifically, the averaged reward at time $ t $ is given by
	\vspace*{-0.5\baselineskip}
	\begin{equation}\label{key}
	\bar{r}(t) = \frac{1}{T_A} \sum_{i = t-T_A+1}^{t} r(i),
	\end{equation}
	where $ T_A $ is a fixed subset size, set to be $ 200 $ in this paper.
	
	\begin{figure*}[tb]
		\centering
		\subfigure[Normalized performance gain]
		{ 	
			\label{fig:cluster_bar}
			\includegraphics[trim = 1 1 1 1, clip, width=8cm]{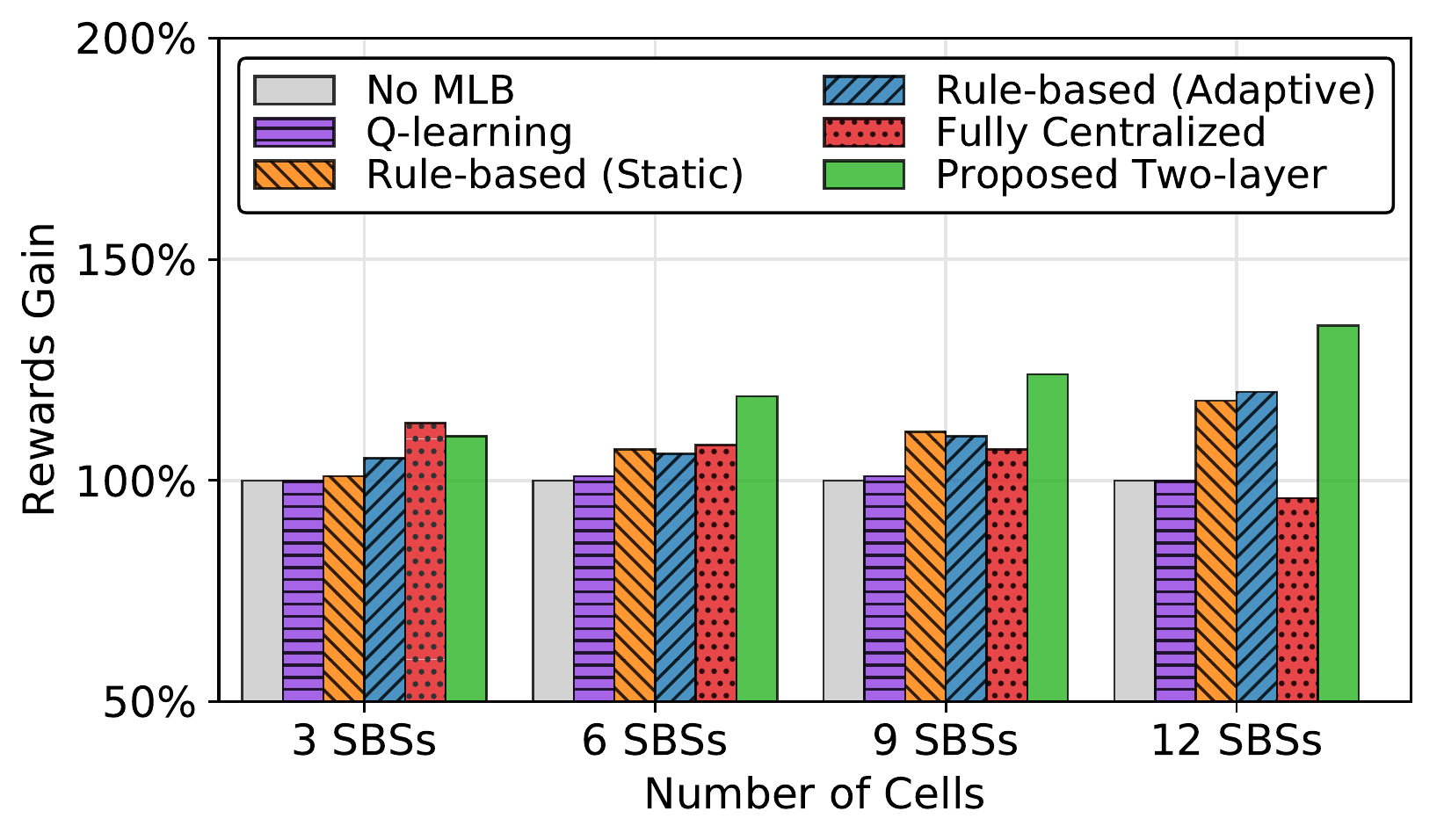}
		}\hspace{5mm}
		\subfigure[Averaged rewards with six SBSs]
		{ 	
			\label{fig:cluster_rewards}
			\includegraphics[trim = 1 1 1 1, clip, width=8cm]{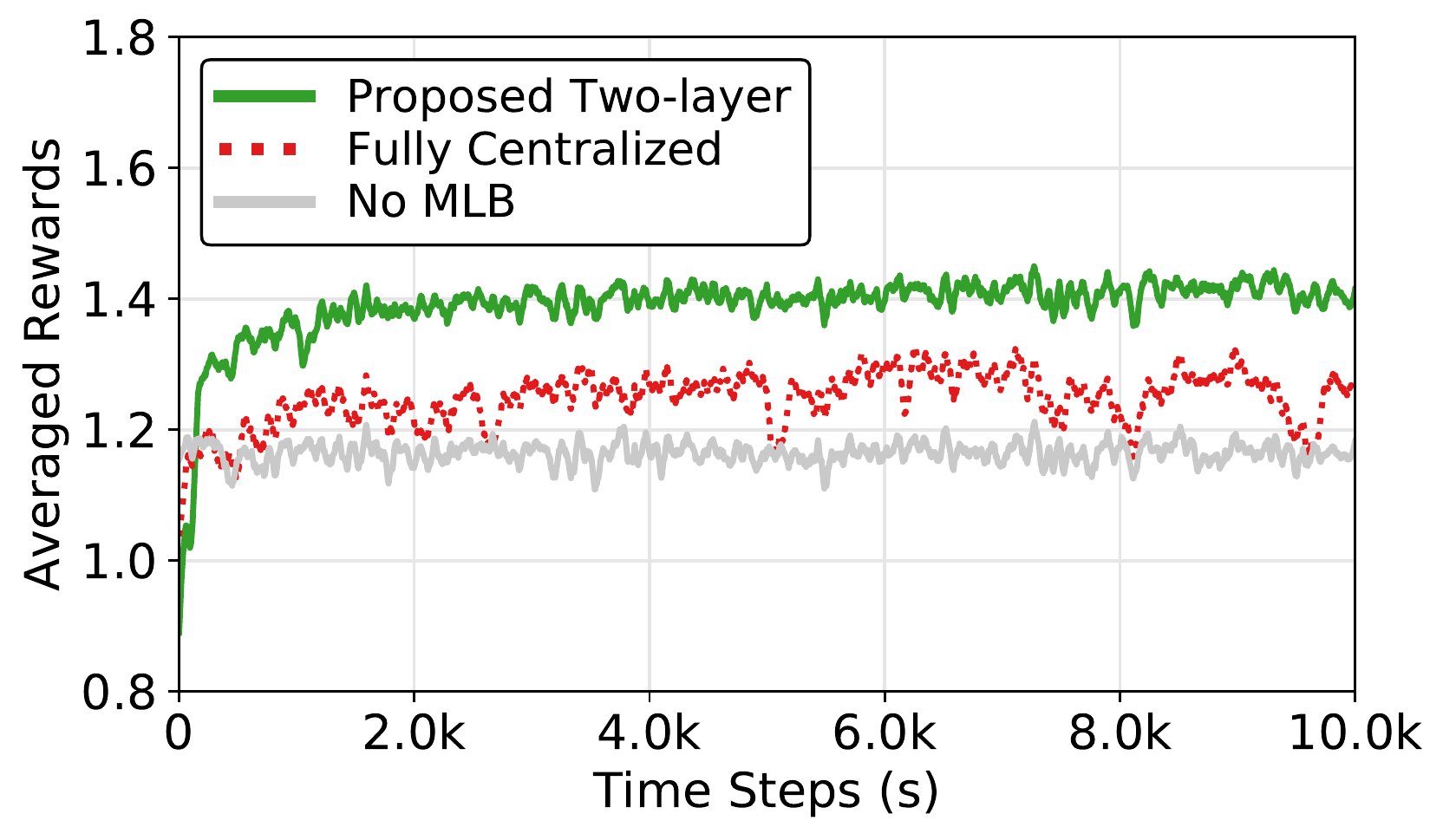}
		}
		\caption{Performance under different number of SBSs}
		\label{fig:cluster}
	\end{figure*}
	
	The result in Fig.~\ref{fig:cluster} presents the scalability of the comparing schemes under the CBR of $ 96~kbps$. The result is averaged over $ 10,000 $ time steps under $ 30 $ different randomized SBS layouts to present a fair and general comparison. 
	Fig.~\ref{fig:cluster_bar} presents the performance gain by using the noMLB method as the reward baseline. The result shows that that when performing MLB among only three SBSs, the centralized architecture generates a better performance due to global optimization. However, when performing MLB among more than three SBSs, the two-layer architecture outperforms the centralized architecture with a growing superiority. This verifies
	the scalability of our proposed two-layer architecture, and validates that the self-organized MLB is promising to find a better (local optimal) solution than the fully centralized MLB by breaking the large non-convex MLB problem into smaller pieces that are easier to handle, as discussed in Sec.~\ref{sec:Two-layer}.
	Meanwhile, the rule-based (static) and the rule-based (adaptive) algorithms are also scalable since they are both fully distributed algorithms. However, their performance is worse than our proposed algorithm due to the lack of cooperations.
	Fig.~\ref{fig:cluster_rewards} presents the detailed averaged rewards obtained with six cells, which reveals that compared with the centralized MLB, the performance of the self-organized MLB is more stable and can converge remarkably faster at the early stage.
	
	\begin{figure}[tb]
		\centering
		\includegraphics[trim = 1 1 1 1, clip, width=8cm]{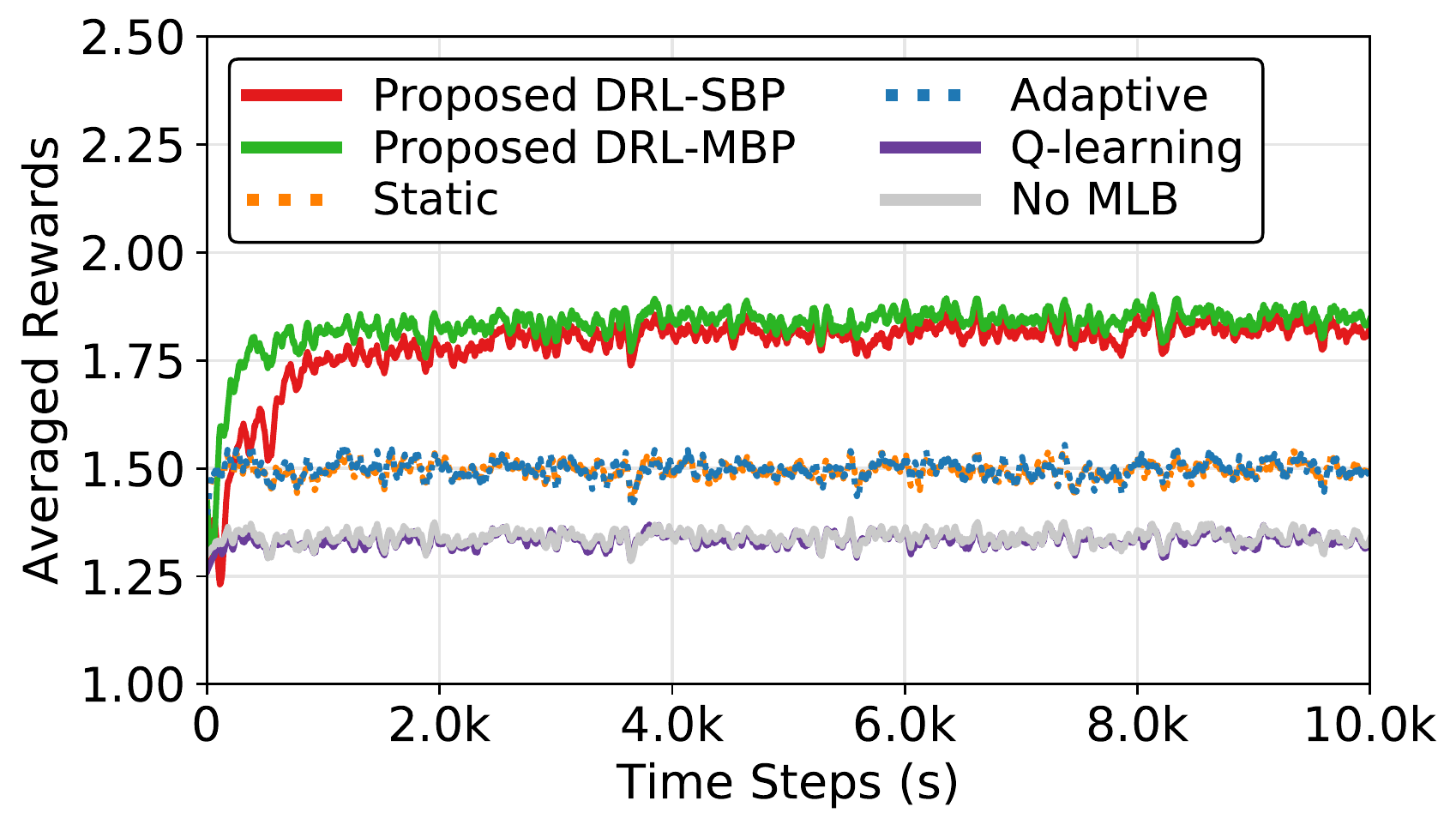}
		\caption{Performance of learning under multiple behavior policies}
		\label{fig:parallel}
	\end{figure}
	The result in Fig.~\ref{fig:parallel} shows the moving averaged rewards of all the comparing schemes over $ 4,000 $ time steps under a CBR requirement of $ 112 $ kbps. The performance is averaged over $ 30 $ different and random SBS topologies. The result shows that the proposed DRL-based algorithms can outperform the competitors considerably. 
	Specifically, for our proposed DRL-based algorithm, we study two alternatives: 1) using a single behavior policy that is constructed by adding certain random noises to the target policy; 2) using three behavior policies, including the above noisy policy and two rule-based policies. We refer to them as the \textit{DRL-SBP} algorithm and the \textit{DRL-MBP} algorithm, respectively. The DRL-SBP algorithm is trained centrally, while the DRL-MBP algorithm is trained in parallel. 
	The noMLB baseline in Fig.~\ref{fig:parallel} indicates that the maximum SBS load fluctuates around 74\% if no MLB attempts are made.
	For the rule-based algorithms, both of the rule-based (static) and rule-based (adaptive) algorithms can control the maximum SBS load to be under 67\%. However, the rule-based (adaptive) algorithm can perform slightly better than the rule-based (static) one.
	For the Q-learning algorithm, due to the quantization of states and actions, the learned policy can hardly be applied to general UDN environments, which results in the same performance as the noMLB baseline.
	For our proposed DRL-based algorithm, both of the DRL-SBP algorithm and the DRL-MBP algorithm can control the maximum SBS load to be under 55\%, where the DRL-MBP algorithm can converge remarkably faster at the early stage and performs slightly better after the convergence.

	\begin{figure*}[tb]
		\centering
		\subfigure[Reward Without Safeguard]
		{ 	
			\label{fig:nosafeguard_rewards}
			\includegraphics[trim =1 1 1 1, clip, width=8cm]{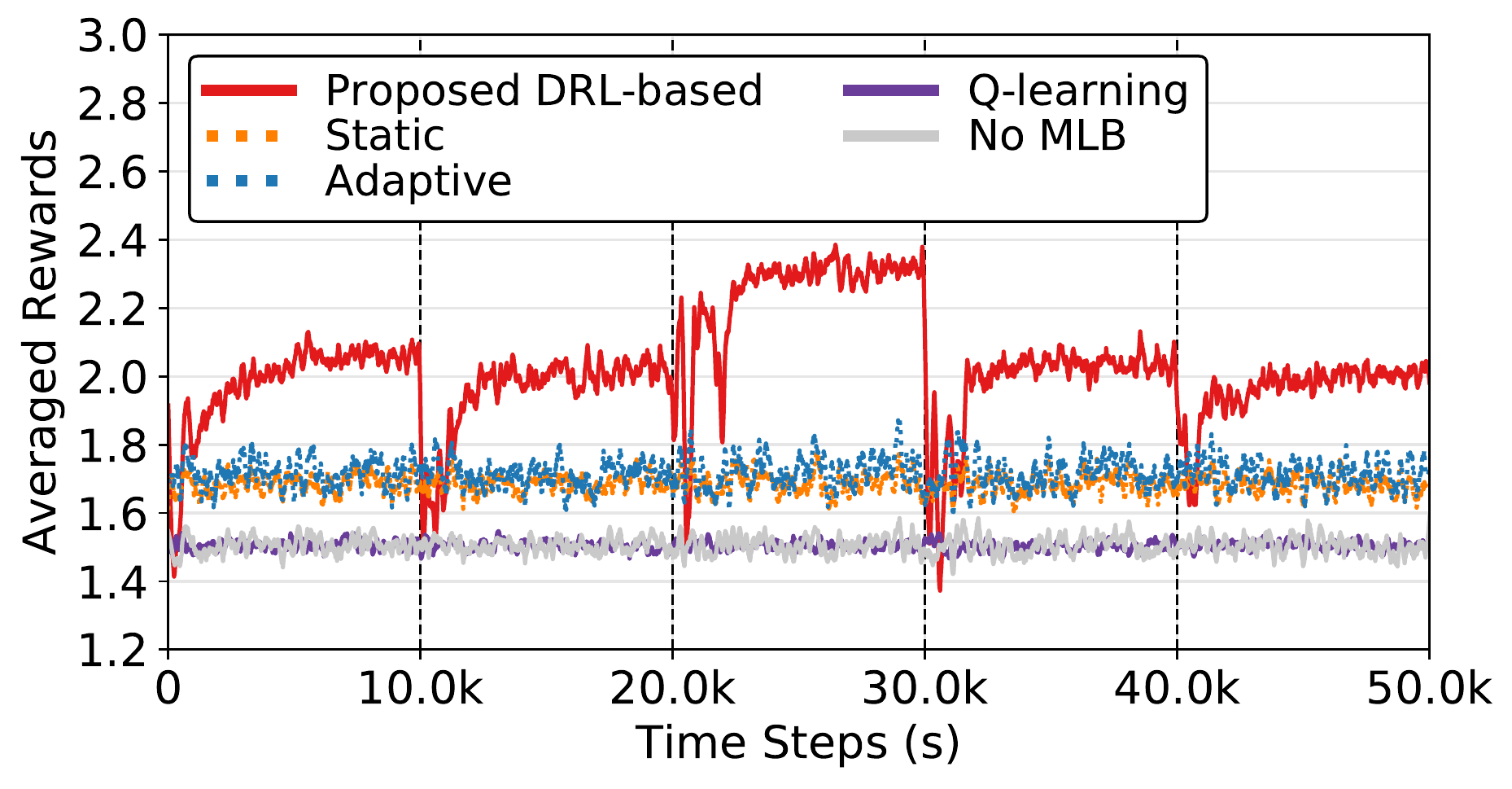}
		}
		\subfigure[Reward With Safeguard]
		{ 	
			\label{fig:safeguard_rewards}
			\includegraphics[trim = 1 1 1 1, clip, width=8cm]{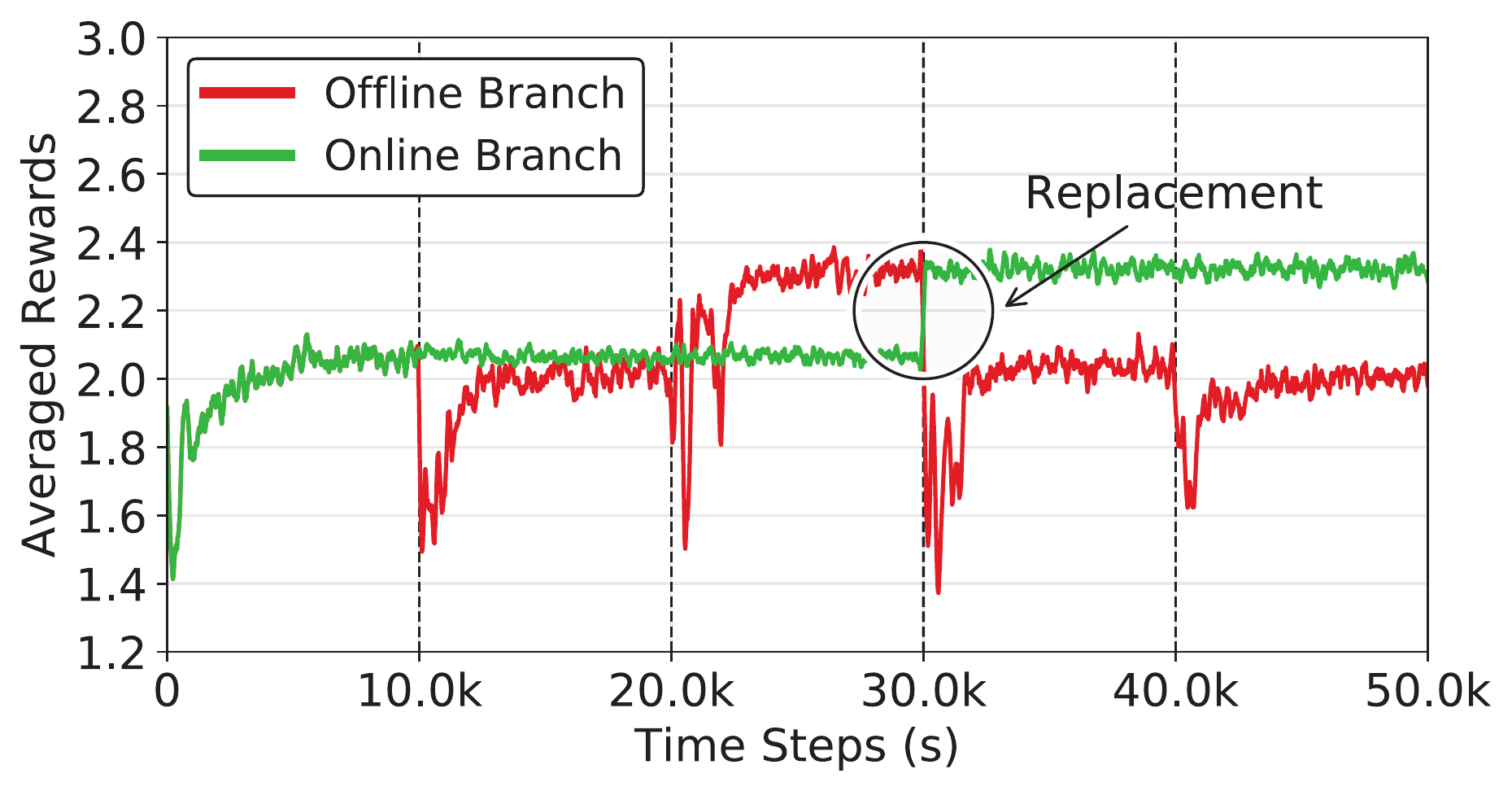}
		}
		\caption{Performance of the safeguard mechanism}
		\label{fig:safeguard_result}
	\end{figure*}
	The result in Fig.~\ref{fig:safeguard_result} shows the rewards obtained by using the proposed safeguard mechanism. The simulation is performed over $ 50,000 $ time steps, where the re-clustering is simply triggered with a period of $ 10,000 $ time steps, i.e., about $ 3 $ hours. Fig.~\ref{fig:nosafeguard_rewards} shows the performance achieved without using the safeguard mechanism. It is clear that the re-clustering operation has a strong impact over the MLB performance.
	Fig.~\ref{fig:safeguard_rewards} shows the performance achieved by using the safeguard mechanism, which can be interpreted as follows. 
	First, at the beginning of the first stage, the offline branch and online branch are initialized with the same parameters and start to learn.
	Next, at the beginning of the second stage, the offline branch triggers re-clustering, and starts to explore a new MLB policy. However, the new policy is worse than the online policy, which is therefore discarded.
	Then, at the beginning of the third stage, the offline branch starts a new exploration and successfully finds a new policy that is better than the the online policy. Hence, the safeguard replace the online policy with the new policy along with the well-trained parameters at the beginning of the forth stage.
	In this way, the MLB performance of the online branch is ensured to be stable and non-decreasing as presented in Fig.~\ref{fig:safeguard_rewards}.
	
	\begin{figure}[tb]
		\centering
		\subfigure[Handover Failure Rate]
		{ 	
			\label{fig:HFO}
			\includegraphics[trim = 1 1 1 1, clip, width=8cm]{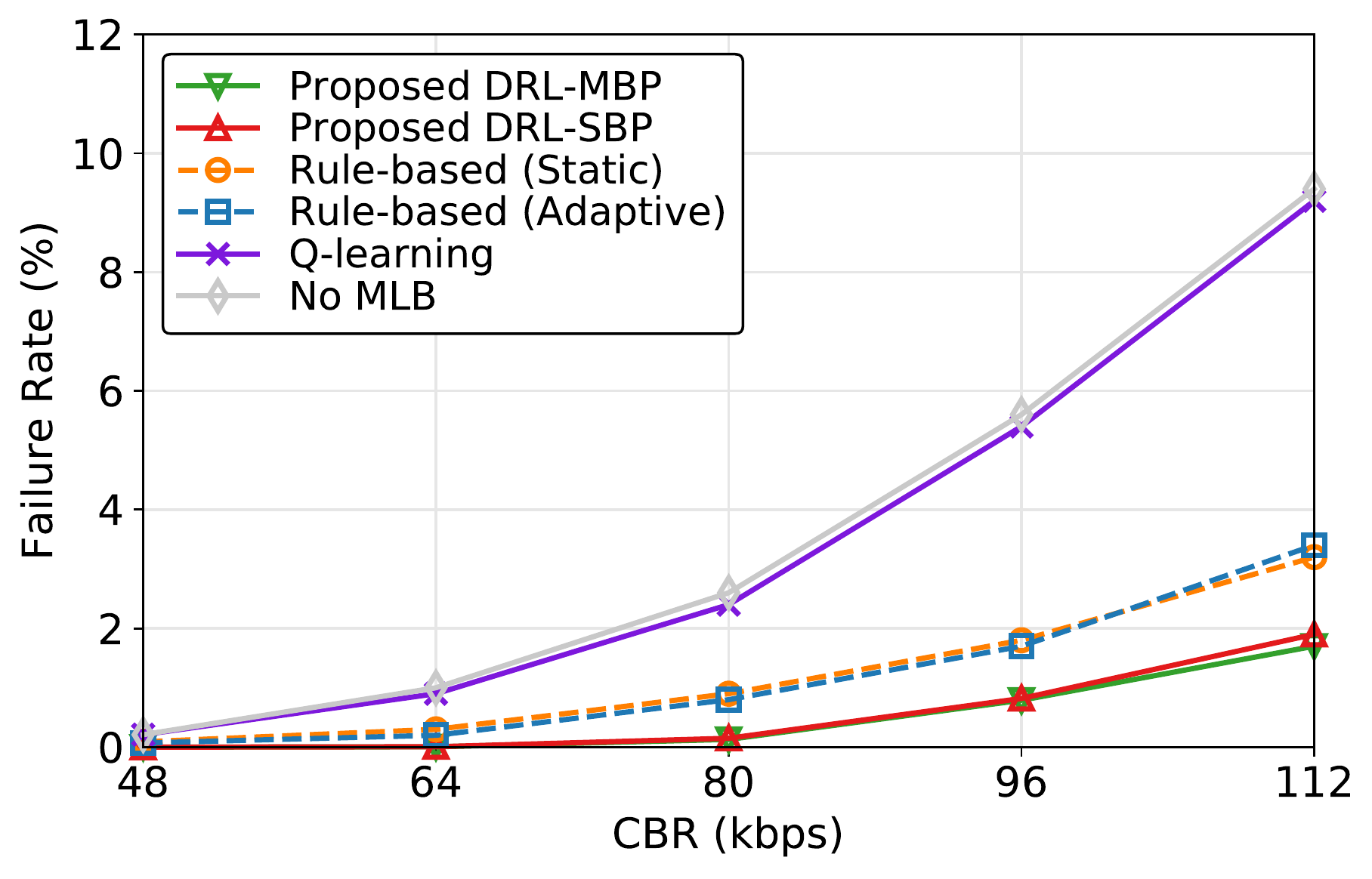}
		}
		\subfigure[Load Standard Deviation]
		{ 	
			\label{fig:std}
			\includegraphics[trim = 1 1 1 1, clip, width=8cm]{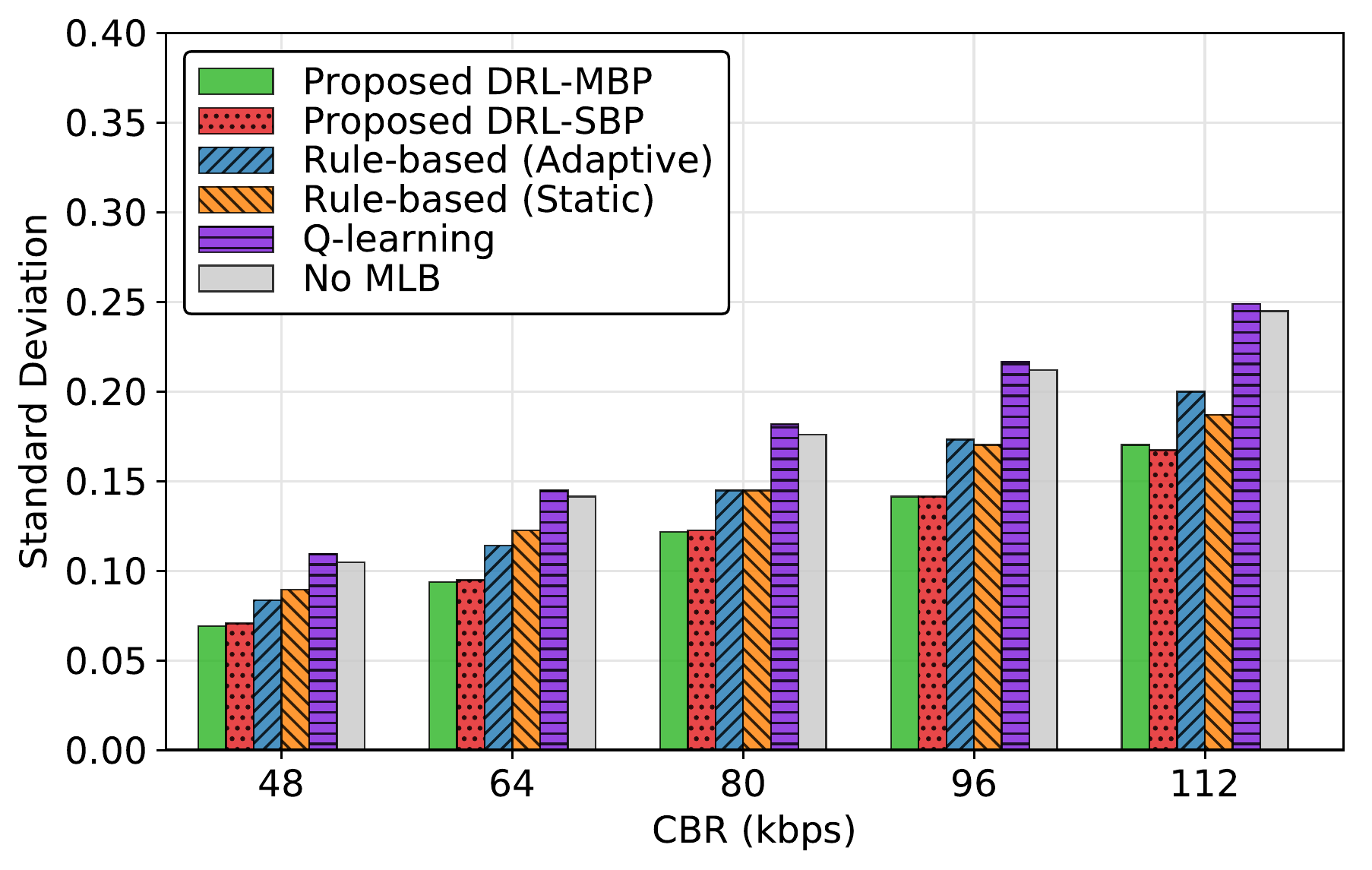}
		}
		\caption{Global load distribution}
		\label{fig:general_performance}
	\end{figure}
	
	The result in Fig.~\ref{fig:general_performance} reflects the global load distribution over the entire area. Specifically, Fig.~\ref{fig:HFO} presents the HFR under different load burdens by varying the CBR from $ 48~kbps $ to $ 112~kbps $. Generally, a lower HFR means that the SBSs are less likely to be overloaded. 
	The result in Fig.~\ref{fig:HFO} shows that the HFO gradually increases with the increase of offered loads for all schemes. However, the proposed DRL-based algorithm can always achieve a lower HFR compared with the other methods. 
	The result in Fig.~\ref{fig:std} shows the load standard deviations under different load burdens. 
	Generally, a lower load standard deviation means that the loads are more equally distributed among SBSs. 
	The result in Fig.~\ref{fig:std} shows that the proposed DRL-based algorithms can achieve low standard deviations as well. Additionally, the superiority of our proposed algorithms increases along with the load burdens.
	The HFR and load standard deviation results jointly verify that our proposed algorithm can achieve a more balanced global load distribution than the competitors.
	
	% ===========================
	%          section
	% ===========================
	\section{Conclusion}
	\label{sec:conclusion}
	In this paper, we proposed a DRL-based MLB algorithm along with a two-layer MLB architecture to solve the large-scale load balancing issue in UDNs. 
	The proposed two-layer architecture can handle the large-scale MLB problem of UDNs in a self-organized manner, which makes it scalable to the number of SBSs.
	The proposed off-policy DRL-based algorithm can be trained via an asynchronous parallel learning framework and employ multiple behavior policies for joint exploration to improve the learning efficiency.
	Moreover, we proposed an off-policy evaluation based safeguard mechanism to improve the online control performance by ensuring that the online UDN system always operate with the optimal and well-trained MLB policy.
	Simulation results verified that 
	1) the proposed algorithm could outperform existing algorithms considerably, in terms of the MLB performance;
	2) the proposed two-layer architecture achieves nice scalability over the number of SBSs, and outperforms the centralized architecture when dealing with a large number of SBSs;
	3) the learning under multiple behavior policies converges remarkably faster than the learning under a single behavior policy;
	4) the safeguard mechanism ensures the online performace to be stable and non-decreasing.
	Moreover, the proposed architecture, algorithm and mechanism are also promising to be applied over other large-scale network control problems in the future networks.
	For example, for the access control problem in the IoT system studied in~\cite{8473693}, one could first cluster the IoT devices into different groups and then determine their access control policies respectively. Meanwhile, the strategy of learning under multiple behavior policies and the safeguard mechanism can be exploited to further improve the learning efficiency and the online performance. More interesting applications can be explored in the future.
	% ==========================================
	% 			References
	% ==========================================
	\bibliographystyle{IEEEtran}
	\bibliography{./bib/IEEEabrv2015,./bib/DRL_Journal}
\end{document}